\documentclass[journal,letterpaper]{IEEEtran}
 \IEEEoverridecommandlockouts

\IEEEoverridecommandlockouts

\def\qed{\rule{1\linewidth}{0.75pt}}
\usepackage{amsfonts}
\usepackage[dvips]{graphicx}
\usepackage{times}
\usepackage{cite}
\usepackage{subfigure}
\usepackage{subfig}
\usepackage{caption}
\usepackage{amsmath,amsthm}
\usepackage{array}
\usepackage{amssymb}

\newcommand{\bs}{\boldsymbol}

\usepackage{stfloats}
\usepackage{graphicx}
\usepackage{footnote}
\usepackage{booktabs}
\usepackage{array}
\usepackage[ruled,linesnumbered]{algorithm2e}
\usepackage{subeqnarray}
\usepackage{cases}
\usepackage{threeparttable}
\usepackage{color}
\usepackage{url}
\usepackage{caption}
\newcommand{\norm}[1]{\left\lVert#1\right\rVert}

\newtheorem{theorem}{Theorem}
\newtheorem{claim}{Claim}
\newtheorem{remark}{Remark}
\newtheorem{proposition}{Proposition}
\newtheorem{lemma}{Lemma}

\newtheorem{definition}{Definition}
\newtheorem{assumption}{Assumption}

\newtheorem{question}{Question}
\newtheorem{corollary}{Corollary}

\usepackage{mathtools}

\def\BibTeX{{\rm B\kern-.05em{\sc i\kern-.025em b}\kern-.08em
    T\kern-.1667em\lower.7ex\hbox{E}\kern-.125emX}}
    
\ifodd 1

\newcommand{\com}[1]{\textbf{\color{red} (Comment: #1) }}
\newcommand{\comg}[1]{\textbf{\color{blue} (COMMENT: #1)}}
\newcommand{\response}[1]{\textbf{\color{blue} (RESPONSE: #1)}}
\else

\newcommand{\com}[1]{}
\newcommand{\comg}[1]{}
\newcommand{\response}[1]{}
\fi

\ifodd 0
\newcommand{\arx}[1]{#1}
\newcommand{\arxv}[1]{}
\else
\newcommand{\arx}[1]{}
\newcommand{\arxv}[1]{#1}
\fi

\begin{document}
\title{Faithful Edge Federated Learning: \\Scalability and Privacy}


\author{
 \IEEEauthorblockN{Meng Zhang, \IEEEmembership{Member, IEEE}, Ermin Wei, \IEEEmembership{Member, IEEE},  Randall Berry, \IEEEmembership{Fellow, IEEE}}
  \vspace{-16pt}
     \thanks{ 
     Part of this work has been presented at NetEcon 2021 \cite{NetEcon}.
     
     This work was supported in part by NSF grants CNS-1908807, AST-2037838, and ECCS-2030251.
     
     Meng Zhang is with the Zhejiang University/University of Illinois at Urbana--Champaign Institute, Zhejiang University, Haining 314400, China (e-mail:jackeymzhang@gmail.com). Part of this work was performed while he was at Northwestern University.
     Ermin Wei and Randall Berry are with the Department of Electrical and Computer Engineering, Northwestern University, Evanston, IL 60208 USA (e-mail:  ermin.wei@northwestern.edu; rberry@northwestern.edu).
     }
 }  
 






\maketitle

\begin{abstract}
Federated learning enables machine learning algorithms to be trained over decentralized edge devices without requiring the exchange of local datasets. Successfully deploying federated learning requires ensuring that agents (e.g., mobile devices) faithfully execute the intended algorithm, which has been largely overlooked in the literature. In this study, we first use risk bounds to analyze  how the key feature of federated learning, unbalanced  and non-i.i.d. data, affects agents' incentives to voluntarily participate and obediently follow traditional federated learning algorithms. To be more specific, our analysis reveals that agents with less typical data distributions and relatively more samples are more likely to opt out of or tamper with federated learning algorithms. To this end, we formulate the first faithful implementation problem of federated learning and design two faithful federated learning mechanisms which  satisfy economic properties, scalability, and privacy. First, we design a \textit{Faithful Federated Learning (FFL) mechanism} which approximates the Vickrey–Clarke–Groves (VCG) payments via an incremental computation. We show that it achieves (probably approximate) optimality, faithful implementation, voluntary participation, and some other economic properties (such as budget balance). Further, the time complexity in the number of agents $K$ is $\mathcal{O}(\log(K))$. Second, by partitioning agents into several clusters, we present a scalable VCG mechanism approximation. We further design a scalable and \textit{Differentially Private FFL (DP-FFL) mechanism}, the first differentially private faithful mechanism, that maintains the economic properties. Our DP-FFL mechanism enables one to make three-way performance tradeoffs among privacy, the iterations needed, and payment accuracy loss.
\end{abstract}


\section{Introduction}\label{Sec:Intro}
\subsection{Motivation}
Machine learning applications often rely on cloud-based datacenters to collect and process the vast amount of needed training data.
Due to the proliferation of Internet-of-Things (IoT) applications, much of this data is generated by devices
in wireless edge networks.
In addition, relatively slow growth in network bandwidth, high latency, and data privacy concerns may make it infeasible or undesirable to upload all the data to a remote cloud, leading some to project that $90\%$ of the global data will be stored and processed locally \cite{zetta}. 
\textit{Federated learning} is a nascent solution to retain data in wireless edge networks and perform machine learning training distributively across end-user devices and edge servers (also called edge clouds) (e.g., \cite{FL1,FL2,FL3,FL4,SDP,Privacy,FL5,FL6,FL7,FL8,FL9,FL91,FL92,FL10,Edge1,Edge2,Edge3,Edge4,Edge5,Edge6,Edge7,Edge8,Edge9}).

Federated learning algorithms aim to fit models to data generated by multiple distributed devices. The large-scale deployment of federated learning relies on overcoming  the following two main technical challenges: the \textit{statistical challenge} and the \textit{communication challenge} \cite{FL1,FL2,FL3}. Specifically, each agent (the owner of each device) generates data in a  non-independently and identically distributed (non-i.i.d.) manner, with the dataset on each device being generated by a distinct distribution and the local dataset size varying greatly. 
Second, communication is often a significant bottleneck in a federated learning framework, motivating  the design of communication-efficient federated learning algorithms. These have been motivating extensive studies on improving efficiency by designing new training models, fast algorithms and quantization techniques (e.g., \cite{FL6,FL7}).
Other studies have been designing algorithms for resource allocation, mobile user selection, energy efficient, scheduling, and new communication techniques  in wireless edge networks (e.g., \cite{Edge1,Edge2,Edge3,Edge4,Edge5,Edge6,Edge7,Edge8,Edge9}).


Nevertheless, whereas many existing federated learning algorithms assume that agents are obedient, i.e., they are willing to follow the algorithms (e.g., \cite{FL1,FL2,FL3,FL4,SDP,Privacy,FL5,FL6,FL7,FL8,FL9,FL10,Edge1,Edge2,Edge3,Edge4,Edge5,Edge6,Edge7,Edge8,Edge9}), edge devices in practice may be \textit{strategic} and may tamper with or opt out of federated learning algorithms to their own advantages. Such strategic edge devices would be more likely to arise in a wireless setting, where different devices may connect and participate in federated learning at different times. 
Therefore, the success of deploying federated learning relies on strategic agents' \textit{voluntary participation (into the federation)} and \textit{faithful} execution of distributed federated learning algorithms. 
Generally speaking, there are two key factors 
that may incentivize  agents to strategically manipulate federated learning:
\begin{itemize}
    \item federated learning may incur significant resource consumption  (e.g., energy, bandwidth, and time) for mobile devices;
    \item agents have different preferences over prediction outcomes (due to, e.g., non-i.i.d and unbalanced data).
\end{itemize}
The above issues reflect an agent's dual role as \textit{a contributor} and \textit{a client} in a federated learning setting, respectively. That is, agents demand for both \textit{rewards for contributing} and \textit{preferred prediction outcomes}.
This work focuses on the latter issue that has been overlooked in the literature, whereas existing studies mainly have been attempting to solve the former one (see the survey in \cite{FLEconSurvey} and references \cite{FLEcon01,FLEcon02,FLEcon03,FLEcon04, FLEcon,FLEcon11,FLEcon12, FLEcon13,FLEcon14, FLEcon3,FLEcon4,FLEcon6, FLEcon7}) by designing incentive mechanisms to reward agents according to agents' data quality, quantity, and reputation.
Specifically,
non-i.i.d. and unbalanced data may render  different prediction objectives for different individual agents and hence may incentivize strategic manipulation. In this case,
each strategic agent can choose either to opt out of or to tamper with the federated learning algorithms to its own advantage.
Such a behavior may result in the failure of large-scale deployments of edge federated learning. To this end,
this paper will first answer the  following question:
\begin{question}
How do unbalanced data and non-i.i.d. data distributions disincentivize agents to obediently follow  and voluntarily participate into federated learning algorithms?
\end{question}


To overcome this issue of manipulation, one approach for the server  is to leverage \textit{(economic) mechanism design}, by anticipating agents' strategic behaviors. As a seminal example, the Vickrey–Clarke–Groves (VCG) mechanism \cite{VCG}
is a generic truthful mechanism for achieving a socially-optimal solution, while ensuring agents' voluntary participation and incentive compatibility, i.e., truthful reports of their local information. However, many such mechanisms use a central authority that computes the optimal solution, which is not applicable in the framework of federated learning. 
To achieve distributed implementation of 
mechanisms, existing studies have developed \textit{faithful} mechanisms (e.g. \cite{Faithful1,Faithful2,Faithful3,Faithful4}), which prevent agents from deviating from the intended algorithms (e.g., by manipulating computation or information reporting).

We note that existing studies on centralized and distributed mechanisms, however,  have not addressed 
the problem of federated learning due to the following important considerations:
\begin{itemize}
   \item \textbf{Scalability}: 
    Mobile devices are expected to be \textit{massively distributed} (i.e., the number of agents may be much larger than the average samples per agent) and have \textit{limited communication} capability \cite{FL1,FL2,FL3}.
    However, existing VCG-based approaches \cite{VCG} involve solving $K+1$ optimization problems (where $K$ is the number of agents), which makes them impractical for large-scale systems.
    \item \textbf{Unknown data distributions}: Existing mechanisms assume that agents know their exact objectives, whereas in federated learning agents' expected objectives are unknown to themselves, since their underlying data distributions are unknown.
    \item \textbf{Privacy}: Federated learning often involves training predictive models based on individuals' private local datasets that contain highly sensitive information (e.g., medical records and web browsing history). For instance, multiple hospitals forecast cancer risks by performing federated learning over the whole patient population, while privacy laws prohibit sharing private patient data \cite{Privacy}. However, existing economic mechanisms require strategic agents to reveal their objective values, which may violate such privacy requirements.\footnote{Specifically, Roberts' theorem states that, under mild conditions, the only incentive compatible mechanisms are VCG variants, which requires the revelation of agents' private information of their loss functions \cite{VCG}. }
 
\end{itemize}


These motivate the following key question:
\begin{question}
	How should one design a \textit{faithful} federated learning mechanism that also achieves \textit{voluntary participation, scalability, and privacy preservation}?
\end{question}

\subsection{Our Work}
In light of the challenges above, this paper studies mechanism design for two representative edge federated learning  scenarios aiming at achieving scalable, privacy-preserving, and faithful edge federated learning. Similar ideas could be applied to other federated learning algorithms.\footnote{We note that designing federated learning algorithms with state-of-the-art performances (e.g., convergence speed, cost-effectiveness, privacy, and robustness) is beyond the scope of  this work.}
We summarize our key contributions in the following:
\begin{itemize}
    \item \textbf{Analysis of risk bounds.} We  analyze how the key feature in federated learning, non-i.i.d. and unbalanced data, affects agents' incentives to voluntarily participate and obediently follow the algorithms. Specifically, our analysis reveals that an agent with a less typical data distribution and relatively more data samples tends to have a greater incentive to opt out of or tamper with federated learning algorithms.
    
    \item  \textbf{Faithful federated learning.} \textit{We design the  first faithful mechanism for federated learning}. It approximates the VCG mechanism and  achieves (probably approximate) optimality,  faithful implementation, voluntary participation, and some other economic properties (such as budget balance). Further, the time complexity in the number of agents $K$ is $\mathcal{O}(\log(K))$.
    
    \item \textbf{Differentially private faithful federated learning.} By partitioning agents into several clusters, we present a scalable VCG mechanism approximation with square root iteration complexity. Based on it, we further design a  Differentially-Private Faithful Federated Learning (DP-FFL) mechanism that is scalable while maintaining VCG's economic properties. 
    In addition, our DP-FFL mechanism enables one to make three-way performance tradeoffs among privacy, convergence, and payment accuracy loss. \textit{To the best of our knowledge, this is the first differentially private and faithful mechanism.}
\end{itemize}

\section{Literature Review} \label{Sec:Related}

\textbf{Federated Learning.} 
 The existing literature has studied how to make the model sharing process more  privacy-preserving (e.g., \cite{FL4,FL5,SDP,Privacy}), more secure (e.g., \cite{FL91,FL92}), more efficient (e.g., \cite{FL6,FL7}), and more robust (e.g., \cite{FL8,FL9}) against heterogeneity in the distributed data sources among many other works. For a more detailed survey, please refer to \cite{FL10}. 
On the other hand, extensive studies attempting to improve the efficiency can be categorized into two directions: \textit{algorithmic} and \textit{communication} design. First, by designing new techniques including quantization (e.g., \cite{FL6}) and new novel learning models (e.g., multi-task federated learning \cite{FL7}). Second, in wireless edge networks, efforts have studied  resource allocation algorithms for edge nodes (e.g., \cite{Edge2}), scheduling policies against interference (e.g., \cite{Edge3,Edge9}), mobile user selection and resource allocation algorithms (e.g., \cite{Edge5,Edge6})
and new communication techniques (e.g., over-the-air computation \cite{Edge7}).
 \textit{However, this line of work assumes that agents (in addition to malicious attackers as in \cite{FL91,FL92}) are willing to participate into federated learning and obey the algorithms,
whereas agents in practice are strategic and require proper incentives to do so.}

In terms of incentive design for federated learning, which has been listed as an outstanding problem in \cite{FL3},
 only a few recent studies attempted to address this issue \cite{FLEcon01,FLEcon02,FLEcon03,FLEcon04, FLEcon,FLEcon11,FLEcon12, FLEcon13,FLEcon14, FL3,FLEcon3,FLEcon4,FLEcon6, FLEcon7}. 
Reference \cite{FLEcon3} describes a payoff sharing algorithm that maximizes the system designer's utility without considering agents' strategic  behaviors.
Yu \textit{et al.} in \cite{FLEcon4} introduced fairness guarantees to the previous reward system. 
Other studies have been considering economic approaches to compensating agents' communication and computation costs based on economic approaches such as contract theory (e.g., \cite{FLEcon6,FLEcon,FLEcon02}), Stackelberg game \cite{FLEcon14,FLEcon11}, auction theory (e.g., \cite{FLEcon12,FLEcon03}), and reputation \cite{FLEcon}.   \textit{However, this line of work focused on incentivizing agent participation by compensating them for their costs and eliciting their truthful cost information, but assumed that
agents are obedient to follow federated learning algorithms without strategic manipulation.} 




\textbf{Faithful Mechanisms.} Only a few studies in the literature considered faithful mechanism design (e.g., \cite{Faithful1,Faithful2,Faithful3,Faithful4}).
Faithful implementation was first introduced by Parkes \textit{et al.} in \cite{Faithful1}: a mechanism is faithful if no one can benefit from deviating, including information revelation, computation, and message passing. 
Feigenbaum \textit{et al.} proposed a faithful policy-based inter-domain routing in \cite{Faithful2}. 
Petcu \textit{et al.} in \cite{Faithful3} generalized the above results and achieve faithfulness for general distributed constrained optimization problems.
\textit{However, none of the existing studies on faithful implementation  guaranteed differential privacy or  scalability, or performed risk bound analysis in a (statistical) learning framework.}

\section{System Model and Problem Formulation} \label{Sec:System}

\subsection{System Overview}

In this section, we introduce our federated learning model which aims to fit a global model over data that resides on, and has been generated by, a set $\mathcal{K}\triangleq\{k: 1\leq k\leq K\}$ of agents (with distributed edge devices).
The model also consists of a trusted (parameter) server.
Each agent (e.g., a mobile device) has access to  a local dataset $\mathcal{D}_k=\{(\bs{x}_i,y_i)\}_{i=1}^{n_k}$, where $\bs{x}_i\in\mathcal{X}\subset\mathbb{R}^{d}$, $y_i\in\mathcal{Y}\subset\mathbb{R}$, $n_k=|\mathcal{D}_k|$ is the number of agent $k$'s  data samples, and $|\cdot|$ denotes the cardinality of a set. Sets $\mathcal{X}$ and $\mathcal{Y}$ are compact. We use $n\triangleq\sum_{k\in\mathcal{K}}n_k$ to denote the total number of data samples and $\mathcal{D}\triangleq \bigcup_{k\in\mathcal{K}} \mathcal{D}_k$ to denote the global dataset.
The training data across the agents are often non-i.i.d., since the data of a given client is typically based on the usage of the particular edge device and may not be representative of the population distribution (e.g., \cite{FL1}).
To model the non-i.i.d. nature of the data, we assume that,
every agent $k\in\mathcal{K}$ generates data via a distinct distribution $P_k(\bs{x},y)$. 



\subsection{Federated Learning Setup}



\subsubsection{Expected risk} Ideally, a federated learning problem fits a global model $\bs{w}\in \mathbb R^d$ via \textit{expected risk minimization}, i.e.,  by minimizing the following (weighted average)
expected risk:
\begin{align}
  {E}(\bs{w})= \sum_{k\in\mathcal{K}}p_k{E}_{k}(\bs{w}), \label{Expected}
\end{align}
where $p_k\geq 0$ represents the weight for each agent $k$, satisfying $\sum_{k\in\mathcal{K}}p_k=1$,\footnote{As an example, \textbf{FedAvg} in \cite{FL1} selects $p_k=n_k/n$ for all $k\in\mathcal{K}$.} and ${E}_{k}(\bs{w})$ is agent $k$'s local expected risk:
\begin{align}
  {E}_{k}(\bs{w})\triangleq \int \ell(\bs{w},\bs{x},y) dP_k(\bs{x},y),~\forall k\in\mathcal{K},
\end{align}
where $\ell(\cdot,\cdot,\cdot)$  is a per-sample loss function dependent on the model $\bs{w}$ applied to the input $\bs x_i$ and the label $y_i$. 

\subsubsection{Agent Modeling}
The non-i.i.d. nature of data implies that agents have different $P_k(\cdot)$ and hence
may have heterogeneous prediction objectives ${E}_{k}(\bs{w})$ and different preferences over the prediction outcome $\bs{w}$. 
Note that as the first work considering agent strategic manipulation in federated learning due to heterogeneous prediction objectives, we disregard the impact of resource consumption incurred in federated learning, which was considered in \cite{FLEcon,FLEcon11,FLEcon12, FLEcon13, FLEcon14,FLEcon4,FLEcon6,FLEcon7}.

Note that, under traditional mechanisms that do not account for federated learning
(e.g., \cite{VCG,Faithful1,Faithful2,Faithful3,Faithful4}), agents were assumed to know their exact objectives before participating. However, this is not the case here since their data distributions are unknown to themselves. Therefore, we assume that agents make their decisions based on probably approximate properties  (instead of the deterministic ones in \cite{VCG,Faithful1,Faithful2,Faithful3,Faithful4}) of the mechanisms to be formally introduced later.

\subsubsection{Empirical risk} 

Each agent's local expected risk ${E}_{k}$ is, however, not directly accessible since $P_k(\cdot,\cdot)$ is unknown. To solve \eqref{Expected} approximately, the induction principle of empirical risk minimization suggests to optimize an objective that averages the loss function on the training sets $\{\mathcal{D}_k\}_{k\in\mathcal{K}}$ instead \cite{ERM}. Mathematically, federated learning algorithms aim to solve the following
(\textit{Empirical Risk Minimization (ERM)}) problem \cite{FL1}:
\begin{align}
 {\rm FL}:\quad \min_{\bs{w}}~F(\bs{w})\triangleq\sum_{k\in\mathcal{K}} 
 p_k F_k(\bs{w}),\label{FL}
\end{align}
where $F_k(\bs{w})$ is the agent $k$'s \textit{local empirical risk (loss)}, given by
\begin{align}
    F_k(\bs{w})=\frac{1}{n_k}\sum_{i=1}^{n_k} \ell(\bs{w},\bs{x}_{i},y_i),~\forall k\in\mathcal{K}.
\end{align}
One can anticipate that the optimal solution $\bs{w}^o$ to \eqref{FL} approximates the solution to \eqref{Expected}.\footnote{Throughout this work, we use empirical risk functions (e.g., $\{F_k\}$) in the objectives of federated learning problems and agents' payments, while we use expected risk functions  (e.g., $\{E_k\}$) for risk bound analysis. }
To quantify such a risk bound, we first
adopt the following standard assumptions on the per-sample loss function $\ell(\cdot,\cdot,\cdot)$
throughout this paper (as in, e.g., \cite{Fast,SDP,Privacy}):
\begin{assumption}[$L_g$-Smoothness]\label{Assum1}
The gradients of the per-sample loss function $\nabla_{\bs{w}} \ell(\bs{w},\bs{x},y)$ are well-defined and continuous such that
there exists a constant $L_g$ satisfying
\begin{align}
    \norm{\nabla_{\bs{w}} \ell(\bs{w}_1,\bs{x},y)-  \nabla_{\bs{w}} \ell(\bs{w}_2,\bs{x},y)}_2 \leq L_g \norm{\bs{w}_1-\bs{w}_2}_2,
\end{align}
for any $\bs{w}_1, \bs{w}_2\in\mathbb{R}^d$ and $(\bs{x}, {y})\in(\mathcal{X},\mathcal{Y})$.
\end{assumption}

\begin{assumption}[$\mu$-Strong Convexity]
The per-sample loss function $\ell(\bs{w},\bs{x},y)$  is $\mu$-strongly convex in $\bs{w}$ for all $(\bs{x}, {y})\in(\mathcal{X},\mathcal{Y})$, i.e., for any $\bs{w}_1, \bs{w}_2\in\mathbb{R}^d, (\bs{x}, {y})\in(\mathcal{X},\mathcal{Y})$,
\begin{align}
    \ell(\bs{w}_2,\bs{x},y)\geq~&\ell(\bs{w}_1,\bs{x},y)+\nabla \ell(\bs{w}_1,\bs{x},y)^{T}(\bs{w}_2-\bs{w}_1)\nonumber\\
    &+{\frac {\mu}{2}}\norm{\bs{w}_2-\bs{w}_1}_2^{2}.
\end{align}
\end{assumption}





Typical examples that satisfy these assumptions include ridge regression, $l2$-norm regularized logistic regression, and softmax classifiers. 
We further use $L_{\ell}$ denote the maximal norm of the gradient of the per-sample loss at $\bs{w}^o$:
\begin{align}
    L_{\ell}\triangleq\max_{(\bs{x},y)\in(\mathcal{X},\mathcal{Y})}\norm{\nabla_{\bs{w}}\ell({\bs{w}^o,\bs{x},y})}_2.
\end{align} 
We can now  characterize the risk bound in the following:
\begin{proposition}[Proof in Appendix \ref{AL1}]\label{P1}
 The following inequality is true with a probability of $1-\delta$:
\begin{align}
   E(\bs{w}^o)-\min_{\bs{w}} E(\bs{w})\leq \sum_{k\in\mathcal{K}}\frac{p_k^2}{n_k}\frac{L_{\ell}^2d\log(2d/\delta)}{4\mu}, \label{EqP1}
\end{align}
where $\bs{w}^o$ is the optimal solution to the federated learning problem in \eqref{FL}, $d$ is the dimension of  $\bs{w}$.
\end{proposition}

The proof of Proposition \ref{P1} involves bounding $\nabla_{\bs{w}} E(\bs{w}^o)$ by the Hoeffding's inequality and leveraging the strong convexity of $E(\cdot)$. 
 In the case of the \textbf{FedAvg} algorithm in \cite{FL1}, which selects $p_k={n_k}/{n}$ for all $k\in\mathcal{K}$, the right hand side of \eqref{EqP1} becomes $\frac{L_{\ell}^2d\log(2d/\delta)}{4\mu n}$, which implies that the bound  in this case converges to $0$ as $n\rightarrow \infty$. The rate $\mathcal{O}(1/n)$ and is comparable to the result in \cite{Fast}, while we do not assume boundedness on $\bs{w}$ as \cite{Fast} did.

\subsection{Goals}

The federated learning problem in \eqref{FL} can be solved efficiently in a centralized manner if the server has the access to the global dataset $\mathcal{D}$, which is, however, impractical in the federated learning setting. Hence, as we mentioned in Section \ref{Sec:Intro}, the focus of federated learning algorithms is on distributed learning that achieves \textit{privacy preservation} and \textit{scalability}. Moreover, here we also seek to ensure that agents will not opt out or tamper with the system.  This 
requires a joint design of a federated learning algorithm and
a proper (economic) mechanism.
Specifically, a mechanism aims to achieve the following economic properties:
\begin{itemize}
    \item (E1) \textit{Efficiency}: The optimal  solution  (or its approximation, e.g., in Proposition \ref{P1}) to \eqref{Expected} is achieved.
    \item (E2) \textit{Faithful Implementation}\footnote{Faithful implementation is also known as \textit{incentive compatibility} or \textit{strategyproofness}.}: Every agent does not have the incentive to deviate from the suggested federated learning algorithm.
    \item (E3) \textit{Voluntary Participation}: Every agent should not be worse off by participating into the mechanism.
    \item (E4) (Weak) \textit{Budget Balance (BB)}: The total payment from agents to the server is non-negative, i.e., the server is not required to inject money into the system.
\end{itemize}
As we have mentioned, we aim to achieve properties (E1)-(E4) in a probably approximate (but not deterministic) manner, due to the unavailability of data distributions $\{P_k\}_{k\in\mathcal{K}}$.

\subsection{Mechanism Design}

In order to achieve the above properties (E1)-(E4), 
the server designs an (economic) mechanism $\mathcal{M}=(\mathcal{A}, \mathcal{S}^m,\bs{w}^*, \mathcal{P})$, described in the following: 
\begin{itemize}
       \item \textit{Strategy space} $\mathcal{A}=\prod_{k\in\mathcal{K}} \mathcal{A}_k$: each agent can select a strategy $A_k\in\mathcal{A}_k$, representing the messages (potential misreports) to be submitted to the server in each iteration of a federated  learning algorithm (such as gradient reporting in \textbf{FedAvg} \cite{FL1});
    \item \textit{(Suggested) protocol/algorithm $\mathcal{S}^m=\{s^m_k\}_{k\in\mathcal{K}}$}: the server would like every agent to follow $\mathcal{S}^m$ by playing $A_k=s^m_k$
    (e.g., agents truthfully reporting their gradients in each iteration in \textbf{FedAvg} \cite{FL1});
    \item \textit{Learning updates} $\bs{w}^*: \mathcal{A}\rightarrow \mathbb{R}^{d}$ describes how the algorithm updates the model $\bs{w}$ in each iteration, which depends on  agents' strategies $\bs{A}$;
    \item \textit{Payment rule $\mathcal{P}=\{\mathcal{P}_k\}_{k\in\mathcal{K}}: \mathcal{A}\rightarrow \mathbb{R}^{K}$} describes the payment each agent $k\in\mathcal{K}$ needs to pay to the server, depending on agents' strategies $\bs{A}$.
\end{itemize}


For a given mechanism $\mathcal{M}$, each agent $k$ aims at minimizing its empirical cost, defined next.
\begin{definition}[Overall loss]\label{Def1}
Each agent $k$ has a (quasi-linear) overall loss, defined as
\begin{align}
    J_k(A_k,\bs{A}_{-k})=\mathcal{P}_k(\bs{A})+F_k(\bs{w}^*(\bs{A})),~\forall k\in\mathcal{K}. \label{OL}
\end{align}
\end{definition}
In \eqref{OL}, we assume that each agent's objective is quasi-linear in its monetary loss, which is a standard assumption in economics \cite{VCG}.


 There are two classical choices of the payment rules in the literature:
\begin{itemize}
    \item The (weighted) VCG payment for each agent $k\in\mathcal{K}$ \cite{VCG}:
\begin{align}
    \mathcal{P}_k^{\rm VCG} = \frac{1}{p_k} \left[\sum_{j\neq k} p_jF_j (\bs{w}^o)-\min_{\bs{w}}\sum_{j\neq k} p_jF_j (\bs{w})\right]\label{VCG}.
\end{align}    
The VCG mechanism is known as a  generic truthful mechanism for achieving a socially-optimal solution (E1) and satisfies incentive compatibility and (E3) and (E4). 
Intuitively, the first term in \eqref{VCG} serves to align each agent's objective to the server's so that each agent also aims to minimize the global loss; the second term ensures that each agent does not overpay so as to ensure voluntary participation (E3). Detailed analysis of the VCG mechanism can be found in \cite{VCG}.
Nevertheless, as we have mentioned, the VCG payment  cannot be  directly applied here due to the communication/computation overhead.  \eqref{FL}. Moreover, it assumes that agents know their exact objectives, which is not true here  due to the unavailability of $P_k(\cdot)$. Hence, in this paper, we will design distributed algorithms and corresponding mechanisms that lead to an approximate VCG payment in \eqref{VCG}, to address the above issues and  attain (E2) and (E3).

    \item Another possible payment rule is the
    Shapley value, which achieves other important properties in cooperative game theory \cite{FLEcon04,Shap}, but will not be addressed in this paper. 
\end{itemize}

\section{Federated Learning Failure due to Strategic Manipulation} \label{Sec:Failure}

 In this section, to illustrate the fact that agents to have incentives to opt out of the federated learning framework and manipulate the algorithms,
 we will demonstrate  how non-i.i.d. and unbalanced data incentivizes strategic agents' misbehaviors.
 Specifically, we will analyze the conditions under which a pure federated learning algorithm (i.e. the one that does not leverage any economic mechanism) for optimizing \eqref{FL} does not satisfy (E2) and (E3).

\subsection{Why May Agents Prefer Local Learning?}
In this subsection, we first understand why agents may rather not to participate into federated learning. In particular, for each agent $k\in\mathcal{K}$, we compare the achievable performances of federated learning (when all agents participate into and obey the algorithm, i.e., to solve \eqref{FL})) and  a \textit{local learning algorithm} to independently solve the following local learning problem based on its local dataset $\mathcal{D}_k$:
\begin{align}
    \bs{w}^{L}_k \triangleq\arg\min_{\bs{w}} {F}_{k}(\bs{w}),~\forall k\in\mathcal{K}. \label{Eq-P1}
\end{align}
In contrast, we use $\bs{w}^{o}$ to denote the optimal solution to the federated learning problem in \eqref{FL}:
\begin{align}
    \bs{w}^{o} \triangleq\arg\min_{\bs{w}}\sum_{k\in\mathcal{K}}p_kF_k(\bs{w}).
\end{align}
We start with the following corollary to understand the performance of local learning:
\begin{corollary}[Proof in Appendix \ref{AL1}]\label{C1}
The following inequality is true with a probability of $1-\delta$:
\begin{align}
    E_k(\bs{w}_k^L)-\min_{\bs{w}} E_k(\bs{w})\leq \frac{L_{\ell}^2d\log(2d/\delta)}{4\mu n_k},\forall k\in\mathcal{K}.\label{localbound}
\end{align}
\end{corollary}
The result in Corollary \ref{C1} is interpreted as a special case of  Proposition \ref{P1}, as shown in Appendix \ref{AL1}.
We next  introduce the following result that compares local learning and federated learning: 
\begin{proposition}[Proof in Appendix \ref{AP2}]\label{P2}
        With a probability of $1-\delta$, 
     federated learning in \eqref{FL} leads to a 
        risk bound of: for every agent  $k\in\mathcal{K}$,
        \begin{align}
            E_{k}(\bs{w}^o)-\min_{\bs{w}}E_k(\bs{w})\leq  &\sum_{j\in\mathcal{K}}\frac{p_j^2}{n_j}\frac{L_{\ell}^2d\log(2d/\delta)}{4\mu}
\nonumber\\
&+2\norm{P_k(\cdot)-\sum_{j\in\mathcal{K}}p_j P_j(\cdot)}, \label{FLbound}
        \end{align}
        where $\norm{f(\cdot)}\triangleq\int_t |f(t)|dt$. 
\end{proposition}

The proof of Proposition \ref{P2} uses the envelope theorem \cite{Envelope}.

The first term in the right-hand side of \eqref{FLbound} characterizes the estimation error. The second term stands for the distance between its \textit{data distribution} and the weighted average data distribution $\sum_{j\in \mathcal{K}}p_jP_j(\cdot)$, which characterizes
how ``typical'' agent $k$ is; a larger value of the second term implies that agent $k$ is less typical.
Intuitively, agent $k$ being more typical implies that training samples of other agents are more useful in solving the agent $k$'s prediction problem and hence resulting in a smaller bound in \eqref{FLbound}.

We note that the bounds in \eqref{localbound} and \eqref{FLbound} are upper bounds that do not show which of the actual (expected) risks is worse. However, since agents do not know their exact data distributions but may estimate how typical they are based on some type of side information,
they may rely on comparing these  upper bounds in \eqref{localbound} and \eqref{FLbound} to decide whether to participate into federated learning.
Specifically, suppose that $p_k = n_k/n$ for all $k\in\mathcal{K}$ so that $1/n=\sum_{j\in\mathcal{K}}p_j^2/n_j$. For an agent with many samples so that $n_k/n$ is close to one, then the first term in \eqref{FLbound} will be close to the term in \eqref{localbound}. Moreover, in such a case if the second term is large enough (i.e. the data is less typical), then this risk bound will be larger than that in \eqref{localbound}. 
On the other hand, if an agent $k$ has only a few samples so that $n_k/n$ is small and the second term in \eqref{FLbound} is small enough  (i.e., its data is typical), then this risk bound in \eqref{FLbound} will be smaller than that in \eqref{localbound}. 
Collectively, we make an important observation from Corollary \ref{L1} and Proposition \ref{P2}:
\begin{remark}\label{R1}
The classical federated learning framework in \eqref{FL} disincentivizes non-typical agents with sufficiently large datasets (such that $1/n_k$ is close to $\sum_{j\in\mathcal{K}}p_j^2/n_j$).
\end{remark}
\subsection{Why May Agents Be Untruthful?}
We next understand the incentive for agents to not follow a suggested federated learning algorithm even if they choose to participate.

Consider a heuristic (manipulation) strategy for agent $k$ to amplify its reports (e.g., its gradients in \textbf{FedAvg} \cite{FL1}) by a constant  $\gamma>1$, in each iteration of a federated learning algorithm. When agents other than $k$ are obedient,
the resultant manipulated federated learning algorithm is equivalent to solving the following problem:
\begin{align}
    \bs{w}_{k,\gamma} =\arg\min_{\bs{w}}\left(\sum_{j\neq k}p_jF_j(\bs{w})+\gamma p_k F_k(\bs{w}) \right).\label{MFL}
\end{align}
We can  derive a risk bound for such a manipulation:
\begin{proposition}[Proof in Appendix \ref{AP3}]\label{P3}
         Suppose that agent $k$ amplifies its report of its gradient by a constant coefficient $\gamma$ and agents other than $k$ report their truthful gradients for \textbf{FedAvg}.
         With a probability of $1-\delta$, the following inequality holds:
        \begin{align}
           &E_{k}(\bs{w}_{k,\gamma})-\min_{\bs{w}}E_k(\bs{w})\label{Untruthful}\\
           \leq& \frac{1}{(1+(\gamma-1)p_k)^2}\left(\sum_{j\neq k}\frac{p_j^2}{n_j}+\frac{\gamma^2 p_k^2}{n_k}\right)\frac{L_{\ell}^2d\log(2d/\delta)}{4\mu}\nonumber\\
            ~&+2\norm{P_k(\cdot)-\frac{1}{{1+(\gamma-1)p_k}}\left(\sum_{j\neq k}p_jP_j(\cdot)+\gamma p_k P_k(\cdot)\right)}\nonumber.
        \end{align}
\end{proposition}

Proposition \ref{P3} is a direct application of Proposition \ref{P2}. Note that, \eqref{Untruthful} becomes exactly the same as in \eqref{localbound} as by letting $\gamma$ approach $\infty$, whereas \eqref{Untruthful} becomes exactly the same as in \eqref{FLbound} when $\gamma=1$. This indicates that, as stated in Remark \ref{R1}, non-typical agents with sufficiently large datasets can benefit from choosing a relatively large $\gamma$.\footnote{It also implies that a strategic non-typical agent with a sufficiently large dataset may
manipulate federated learning and lead to a system performance as worse as that of local learning.}
Further, since local learning and federated learning without manipulation can be regarded as the special cases of the manipulated federated learning in \eqref{MFL},
tampering with federated learning renders more capability and incentives to manipulate the system outcomes, 
compared to opting out of federated learning.


   	\begin{figure*}[t]
  	\centering
		\subfigure[]{\includegraphics[scale=.32]{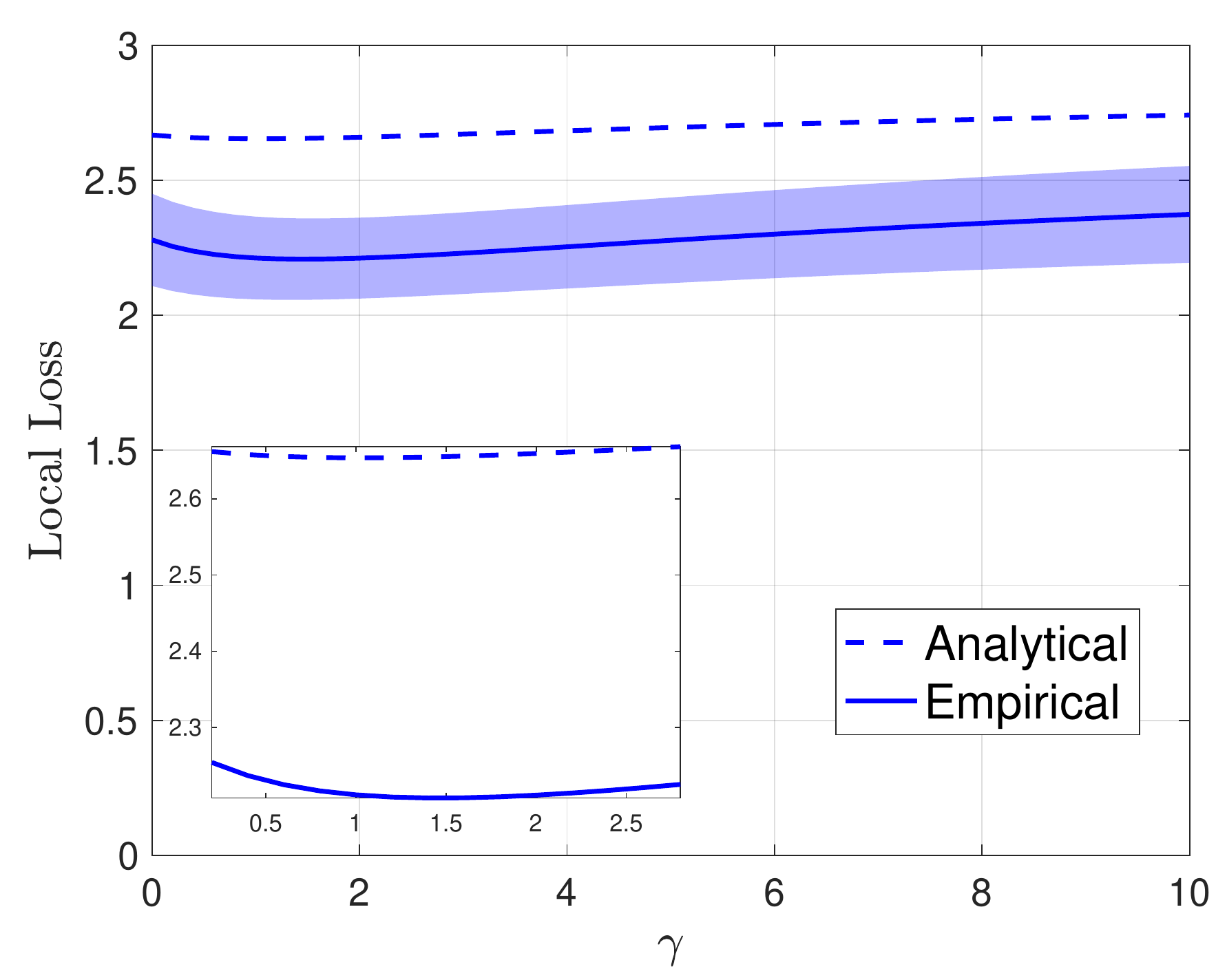}}
		\subfigure[]{\includegraphics[scale=.32]{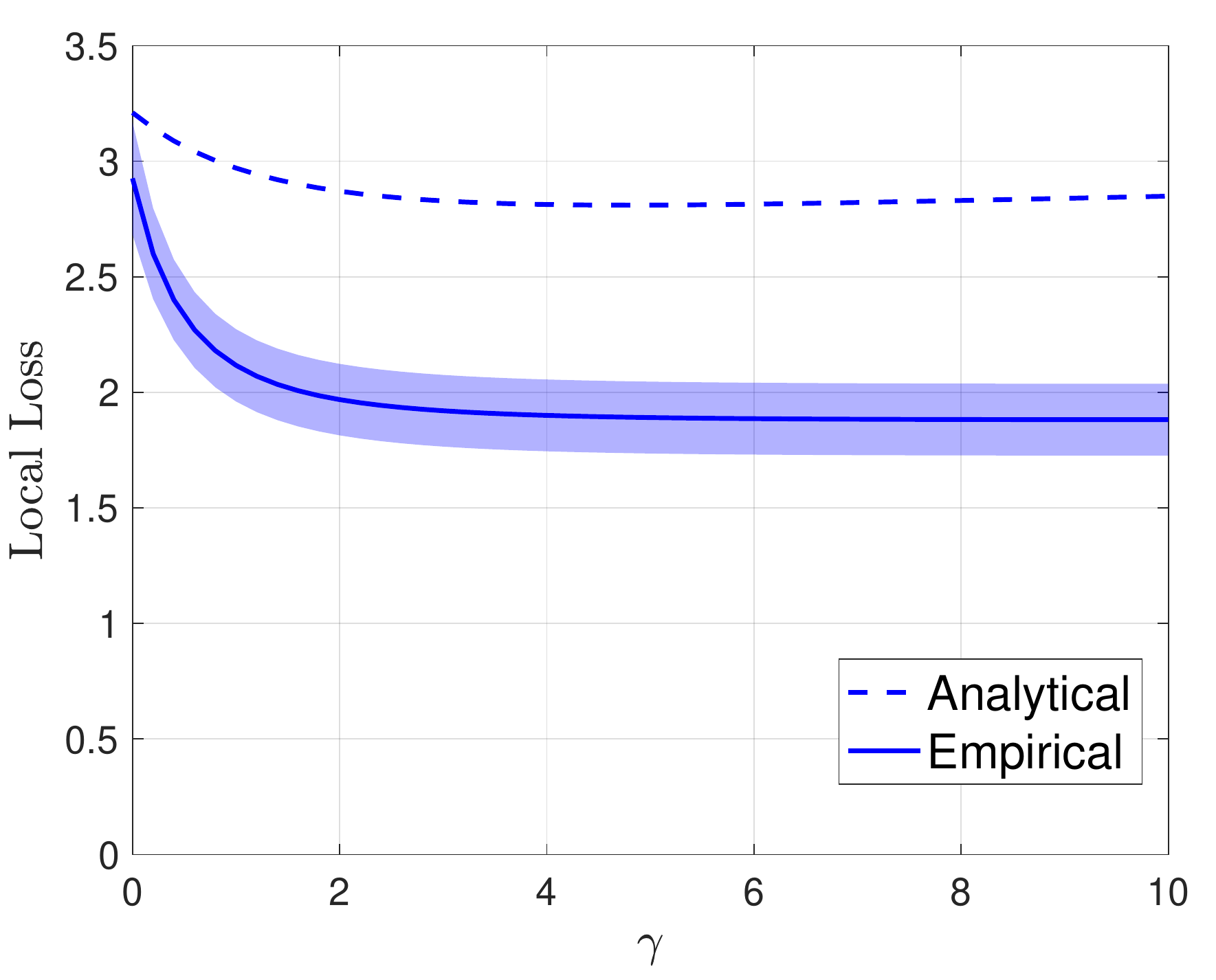}}
		\subfigure[]{\includegraphics[scale=.32]{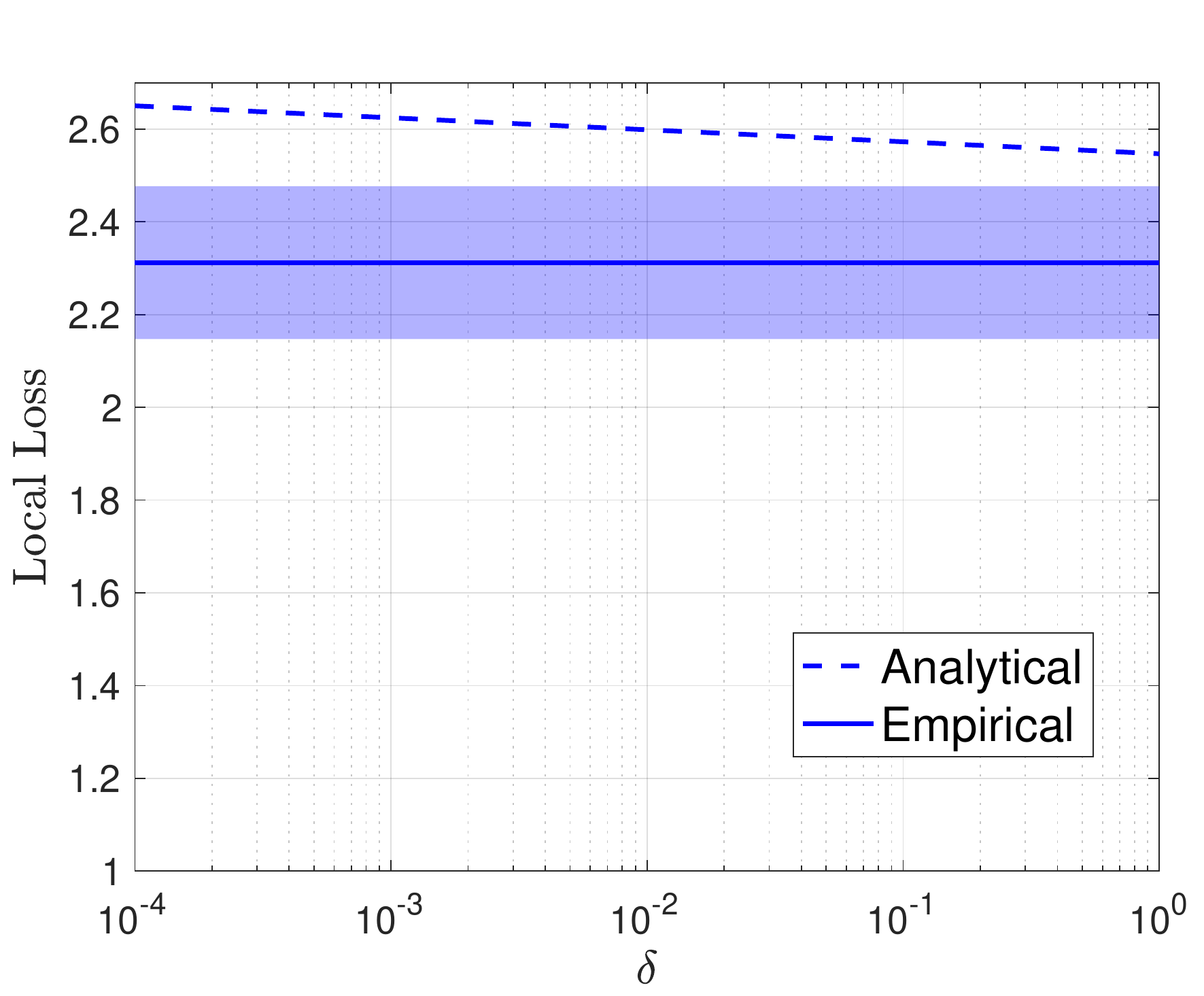}}
		\caption{An illustrative example of Proposition \ref{P3}. In (a) and (b), we  set $\text{mean}=0.1$ and $2$, respectively, fix $\delta=0.01$, and
		compare the incurred local loss at different $\gamma$;  in (c), we fix $\gamma=3$ and $\text{mean}=0.1$ and compare the incurred local loss at different $\delta$. 
 }
 \label{Illustration}
	\end{figure*} 
	
We consider an illustrative example of Proposition \ref{P3} (and Remark 1), as descried in the following. We consider a federate learning framework for 
two agents with $n_1=50$ and $n_2=400$ data samples, respectively. For each data sample $(x_i,y_i)$ in either dataset $\mathcal{D}_1$ and $\mathcal{D}_2$, $x_i\in\mathbb{R}$ is randomly generated from a uniform distribution over $[0,1]$. Labels $y_i$ satisfy $y_i=-2x_i+1+\kappa_i$, where $\kappa_i$ follows normal distributions $\mathcal{N}(\text{mean},2)$ and $\mathcal{N}(0,2)$, truncated over $[-3,3]$, for each data sample $(x_i,y_i)$ generated in datasets $\mathcal{D}_1$ and $\mathcal{D}_2$, respectively. Therefore, a $\text{mean}$ closer to $0$ implies that both agents' data distributions are closer.
In Fig. \ref{Illustration},
we study the impacts of  the amplifying coefficient $\gamma$ and $\delta$
on
agent $1$'s empirical loss 
and its analytical (probably approximate) upper bound derived according to Proposition \ref{P3}. 
We consider a scenario where two data distributions are close (as $\text{mean}=0.1$) in Fig. \ref{Illustration}(a). 
First, we observe that the agent $1$'s optimal coefficients $\gamma$
to minimize its analytical and empirical local risks are both slightly larger than $1$. On the other hand, when two data distributions are very distant (as $\text{mean}=2$) as shown in Fig. \ref{Illustration}(b), agent $1$ prefers an infinitely large $\gamma$. 
Note that each agent's local loss as $\gamma\rightarrow\infty$ corresponds to its local loss under local learning.
By comparing the values of local losses at $\gamma=1$ and $\gamma=\infty$,
Fig. \ref{Illustration} also validates our result in Remark 1: A non-typical agent prefers not to participate.
  Fig. \ref{Illustration}(a)-(b) shows that analytical and empirical loss functions tend to have close minimizers.
 Finally, Fig. \ref{Illustration}(c) demonstrates that $\delta$ only has a small impact on the tightness of the analytical bound since $\delta$ appears in a logarithmic function; $\delta=0.01$ is already enough to ensure a reasonably tight analytical  upper bound.


\subsection{When is Federated Learning Socially Efficient?}

Even from the system-level perspective, with respect to minimizing the global expected risk in \eqref{Expected}, federated learning may not be always more beneficial than local learning in \eqref{Eq-P1}, as we will analyze next.

With a probability of at least $1-\delta$, agents performing their respective local learning algorithms leads to an expected risk bound  of 
\begin{align}
      &\sum_{k\in\mathcal{K}}p_k E_k(\bs{w}_k^L)-  \sum_{k\in\mathcal{K}}p_k\left(\min_{\bs{w}} E_k(\bs{w})\right)\nonumber\\
      \leq & \sum_{k\in\mathcal{K}}\frac{p_k}{n_k}\frac{L_{\ell}^2d\log(2Kd/\delta)}{4\mu}.\label{SE-L}
\end{align}
Similarly, with a probability of at least $1-\delta$, federated learning in \eqref{FL} leads to an expected risk bound of 
\begin{align}
    E(\bs{w}^o)-\sum_{k\in\mathcal{K}}p_k\left(\min_{\bs{w}} E_k(\bs{w})\right)\leq & \sum_{k\in\mathcal{K}}\frac{p_k^2}{n_k}\frac{L_{\ell}^2d\log(2Kd/\delta)}{4\mu}\nonumber\\
& +\sum_{k\in\mathcal{K}}2p_k\norm{P_k-\sum_{j\in\mathcal{K}}p_j P_j}.\label{Eq16}
\end{align}
Consider a system with $n_k=n_j$ and $p_k=p_j$ for all $k$ and $j$. Subtracting the risk bound of local learning from that of federated learning yields
 \begin{align}
     \frac{(K-1)L_{\ell}^2d\log(2Kd/\delta)}{4\mu n}-\frac{2}{K}\sum_{k\in\mathcal{K}}\norm{P_k-\sum_{j\in\mathcal{S}}\frac{P_j}{K}}.\label{subs}
 \end{align}
We observe that the first term in \eqref{subs} always increases in $K$ while the second term needs not to do so, which implies that the federated learning is more likely to have a smaller risk bound when there are many agents in the system.



To generalize the above results, we can further cluster agents into several disjoint clusters $\mathcal{C}_1, \mathcal{C}_2,...,\mathcal{C}_L$, and let agents in each cluster perform \textit{intra-cluster learning}, i.e., they solve
\begin{align}
  \bs{w}^{\mathcal{C}_l}\triangleq\arg\min_{\bs{w}} \sum_{k\in\mathcal{C}_l}p_kF_k(\bs{w}).
\end{align}
Let $\hat{\mathcal{C}}(k)$ denote the cluster that agent $k$ belongs to, i.e., if $k\in\mathcal{C}_{l}$, then $\hat{\mathcal{C}}(k)=\mathcal{C}_l$.
The risk bound of such clustering and intra-cluster learning
is given by
\begin{align}
& \sum_{k\in\mathcal{K}}p_k E_k(\bs{w}^{\hat{\mathcal{S}}(k)})-  \sum_{k\in\mathcal{K}}p_k\left(\min_{\bs{w}} E_k(\bs{w})\right) \nonumber\\
\leq &~\sum_{k\in\mathcal{K}}\frac{p_{j,\hat{\mathcal{C}}(k)}^2}{n_k}\frac{L_{\ell}^2d\log(2Kd/\delta)}{4\mu}
+\sum_{k\in\mathcal{K}}2p_k\norm{P_k-\bar{P}_{\hat{\mathcal{C}}(k)}}\nonumber\\
\triangleq &~{\rm RB}_{\mathcal{C}},\label{Ds}
\end{align}
where $\mathcal{C}$ denotes the set of all clusters, i.e., $\mathcal{C}=\{\mathcal{C}_1, \mathcal{C}_2,...,\mathcal{C}_L\}$, $p_{k,\mathcal{A}}=p_k/\sum_{j\in\mathcal{A}}p_j$, and $\bar{P}_{\mathcal{A}}(\cdot,\cdot)=\sum_{j\in\mathcal{A}}p_{j,\mathcal{A}} P_j(\cdot,\cdot)$, for all clusters $\mathcal{A}$ in $\mathcal{C}$.

It is possible to design a clustering algorithm, based on agents' non-i.i.d. data distributions, to minimize the risk bound, which is beyond the scope of this work. Instead, we focus on the scenario in which federated learning is more beneficial than any clustering $\mathcal{C}$ for the system, by
adopting the following assumption:
\begin{assumption}\label{Assum5}
The federated learning leads to a tighter risk bound than any possible intra-cluster federated learning, i.e.,
\begin{align}
    {\rm RB}_{\{\mathcal{K}\}}\leq {\rm RB}_{\mathcal{C}},~\forall \mathcal{C}.
\end{align}
\end{assumption}
{Assumption \ref{Assum5} can be true in a \textit{massively distributed} federated learning framework \cite{FL1}, in which the average number of samples per agents $n/K$ is much smaller than $K$.}
Even under Assumption \ref{Assum5}, agents may still benefit from opting out of and tampering with a federated learning algorithm, which motivates the faithful federated learning mechanisms to be discussed in the following sections.

\section{Faithful Federated Learning}\label{Sec:FFL}


In this section, we apply mechanism design for \textbf{FedAvg} in \cite{FL2} to achieve a
scalable federated learning algorithm (with the associated mechanism) that satisfies (E1)-(E3) approximately  and (E4) exactly. Similar techniques could also be applied to other federated learning algorithms. 

\begin{algorithm}[t]\caption{Faithful Federated Learning}
\label{Algo1}
\SetAlgoLined

 The server initializes a model $\bs{w}[0]$ and step sizes $\eta_1$ and $\eta_2$ for Phase I and Phase II\;

 \tcp{Phase I: Federated learning phase}\label{l1-Algo1}

 \For{\textbf{iterations} $t\in\{0, 1,..., T_1\}$}{ 
The server broadcasts the current model $\bs{w}[t]$ to all agents\;

Each agent $k$ computes and reports its gradient to the server $\nabla {F}_k(\bs{w}[t])$.\;\label{l4}

The server updates the model
        \begin{align}
       \hspace{-2cm}
       \bs{w}[t+1]=\bs{w}[t]- \eta_1\sum_{k\in\mathcal{K}} p_k\nabla {F}_k(\bs{w}[t]). \label{model-Algo1}\hspace{-2cm}
        \end{align}
 }
\textit{\textbf{Return}} the approximately optimal model: $\bs{w}^*\triangleq\bs{w}[{T}_1]$\; \label{l7-Algo1}
  \tcp{Phase II: Payment computation phase}
 \For{\textbf{agents} $k\in\mathcal{K}$}{
\label{l8-Algo1}
   The server initializes the model $\bs{w}_{-k}[0]=\bs{w}^*$\;
   
 \While{\textbf{iterations} $t\in\{0, 1,..., T_2\}$ or $\frac{1}{2\mu}\norm{\sum_{j\neq k}p_j\nabla F_j(\bs{w}_{-k}[t])}_2^2> p_k \epsilon $\label{l10-Algo1}}  
 { 
 Set $t\rightarrow t+1$\;
 The server broadcasts the current model $\bs{w}_{-k}[t]$ to all agents excluding $k$\;
 
 Agents $j\neq k$ compute and report their gradients to the server $\nabla {F}_j(\bs{w}_{-k}[t])$\;\label{l12}
 
 The server updates the model and agent $k$'s payment:
 \begin{subequations}\label{payment-Algo1}
        \begin{align}
         \bs{w}_{-k}[t+1]&=\bs{w}_{-k}[t]- \eta_2\sum_{j\neq k} p_j\nabla {F}_j(\bs{w}_{-k}[t]),\label{payment-1} \\
    \hspace{-2cm}      \mathcal{P}_k[t+1]&=\mathcal{P}_k[t]+(\bs{w}_{-k}[t+1]-\bs{w}_{-k}[t])^T\cdot  \nonumber\\&~~~~~\sum_{j\neq k}\frac{p_j}{p_k} \nabla {F}_j(\bs{w}_{-k}[t]).\label{payment-2}
        \end{align}        
 \end{subequations}

 }
\textit{\textbf{Return}} the payment for agent $k$: $\mathcal{P}_k^*\triangleq\mathcal{P}_k[T_2]$\; 
 }
 \label{l15-Algo1}
\end{algorithm}

\subsection{Algorithm and Mechanism Description}
We present the \textbf{Faithful Federated Learning (FFL)} algorithm in Algorithm \ref{Algo1}, consisting of two phases: a \textit{federated learning phase} (lines \ref{l1-Algo1}-\ref{l7-Algo1}) and a \textit{payment computation phase} (lines \ref{l8-Algo1}-\ref{l15-Algo1}). 
In the first phase (consisting of $T_1$ iterations), we present a gradient-based federated learning algorithm to (approximately) attain the globally optimal solution $\bs{w}^*$ to \eqref{FL}, similar to FedAvg in \cite{FL1}.\footnote{Throughout this paper, we use superscript $o$ to denote the exact optimal solution, and $*$ to denote the solution output by algorithms.}
The second phase consists of $K$ outer iterations, with each computing each agent's payment without the need of directly revealing agents' private local empirical risk functions. We note that $\frac{1}{2\mu}\norm{\sum_{j\neq k}p_j\nabla F_j(\bs{w}_{-k}[t])}_2^2> p_k \epsilon$ in line \ref{l10-Algo1} is to ensure the accurate computation of each agents payment.
As we will show later, accurately computed payments incentivize strategic agents to faithfully follow the intended federated learning algorithm in the first phase.


Based on Algorithm \ref{Algo1}, we introduce the  FFL mechanism. The reporting of each agent's true gradient  (lines \ref{l4} and \ref{l12}) in each iteration corresponds to the intended algorithm (that the server would like agents to follow), whereas each agent $k$ report $\nabla \tilde{F}_k(\bs{w}[t])\in\mathbb{R}^{d}$ is a potential misreport of its gradient. Formally, we have:
\begin{definition}[The  FFL Mechanism]
The  FFL mechanism $\mathcal{M}=(\mathcal{A}, \mathcal{S}^m,\bs{w}^*, \mathcal{P})$ is described as: for every agent $k\in\mathcal{K},$
\begin{align}
    s^m_k=\{\nabla F_k(\bs{w}[t])\}_{t\in \mathcal{T}_{-k}}\quad{\rm and}\quad  A_k=\left\{\nabla \tilde{F}_k(\bs{w}[t])\right\}_{t\in \mathcal{T}_{-k}}, 
\end{align}
where $\mathcal{T}_{-k}=\bigcup_{j\in\{0\}\cup\mathcal{K}\backslash\{k\}}\{1\leq t\leq T_2\}$. The output global model $\bs{w}^*$ and each agent $k$'s payment $\mathcal{P}_k$ are determined in \eqref{model-Algo1} and \eqref{payment-Algo1}, respectively.
\end{definition}

In the following, we explain the intuition behind Algorithm \ref{Algo1} and the FFL mechanism. We first define
\begin{align}
    \bs{w}_{-k}^o\triangleq\arg \min_{\bs{w}} \sum_{j\neq k}p_jF_{j}(\bs{w}).\label{-k}
\end{align}
Following Algorithm \ref{Algo1}, the final payment for each agent $k$ can be approximately expressed by
\begin{align}
\mathcal{P}_k^*&=\sum_{t=1}^{T_2}\sum_{j\neq k}  \frac{p_j}{p_k} \nabla F_j(\bs{w}[t])^{T} (\bs{w}[t+1]-\bs{w}[t])\nonumber\\
    &\approx \sum_{j\neq k} \int_{\bs{w}_{-k}[0]}^{\bs{w}_{-k}[T_2]}  \frac{p_j}{p_k} \nabla F_j(\bs{w})^T \bs{d} \bs{w}\nonumber\\
    &\approx\sum_{j\neq k}\frac{p_j}{p_k}\left[F_j(\bs{w}^o)-F_j(\bs{w}^o_{-k})\right],~\forall k\in\mathcal{K}.
\end{align}
Therefore,
 the payment rule defined in Algorithm \ref{Algo1} is a VCG-like payment rule (i.e., it approximates \eqref{VCG}). As we have discussed, it can align each agent's objective to the server's objective, and hence potentially satisfies properties (E1)-(E4). 

\subsection{Federated Learning Phase}

With Assumptions 1-3, the gradient-based learning algorithm in the federated learning phase of Algorithm \ref{Algo1} leads to the following standard linear convergence result \cite{conver}:
\begin{lemma}\label{L1}
In the federated learning phase of Algorithm \ref{Algo1}, if we choose a constant step size such that $\eta_1= 1/L_g$, 
the gradient descent has a linear convergence rate of 
\begin{align}
              F(\bs{w}^*)-F(\bs{w}^o)\leq \left(1-\frac{\mu}{L_g}\right)^{T_1} ( F(\bs{w}[0])-F(\bs{w}^o)).
\end{align} 
\end{lemma}
Lemma \ref{L1} along with Proposition \ref{P1} implies that we can achieve (E1) in a \textit{probably approximate} manner, as shown in the following:
\begin{proposition}[Proof in Appendix]\label{P3.5}
In the federated learning phase of Algorithm \ref{Algo1}, if we choose a constant step size such that $\eta_1\leq 1/L_g$, the following \textit{risk bound} is true with a probability of $1-\delta$:
\begin{align}
E(\bs{w}^*)-& \min_{\bs{w}} E(\bs{w})\leq \Phi(\delta),\label{EqP4}
\end{align}
for any $\delta\in(0,1)$, where
\begin{align}
    \Phi(\delta)\triangleq& \sum_{k\in\mathcal{K}}\frac{p_k^2}{n_k}\frac{L_{\ell}^2d\log(2d/\delta)}{2\mu}\nonumber\\
&+\frac{2L_g}{\mu}\left(1-\frac{\mu}{L_g}\right)^{T_1} ( F(\bs{w}[0])-F(\bs{w}^o)). \label{Phi}
\end{align} 
\end{proposition}


\subsection{Payment Computation Phase}
In this subsection, we analyze the payment computation phase.
We start with defining an optimal solution set $\mathcal{W}^*$ for all possible  $\{p_k\}$:
  \begin{align}
  \mathcal{W}^*\triangleq\left\{\arg\min_{\bs{w}} \sum_{k\in\mathcal{K}}p_k F_k(\bs{w}): \forall p_k\geq 0~{\rm and}~\sum_{k\in\mathcal{K}}p_k=1\right\},\label{BigW}
\end{align}
and introduce the following
 gradient bound:
\begin{definition}[Gradient Bound]\label{DefAsum2}
The gradient bound $L_f$ is defined as, for each agent $k\in\mathcal{K}$,
    \begin{align}
  \norm{\nabla_{\bs{w}}F_k(\bs{w})}_2\leq L_f, 
  \end{align}
  for all $\bs{w}\in\{\bs{w}[t]\}_{t\in\{0,1,...,T_1\}}\cup\{\bs{w}_{-k}[t]\}_{k\in\mathcal{K}, t\in\{0,1,...,T_2\}}\cup \mathcal{W}^*$,
  where $\{\bs{w}[t]\}_{t\in\{0,1,...,T_1\}}$ and $\{\bs{w}_{-k}[t]\}_{k\in\mathcal{K}, t\in\{0,1,...,T_2\}}$ are described in Algorithm \ref{Algo1}.
\end{definition}
Note that such a gradient bound always exists 
as the set $\{\bs{w}[t]\}_{t\in\{0,1,...,T_1\}}\cup\{\bs{w}_{-k}[t]\}_{k\in\mathcal{K}, t\in\{0,1,...,T_2\}}\cup  \mathcal{W}^*$ 
is compact\footnote{
By the maximum theorem and the strong convexity of $F_k(\bs{w})$, $\arg\min_{\bs{w}} \sum_{k\in\mathcal{K}}p_k F_k(\bs{w})$ is continuous in $\{p_k\}_{\in\mathcal{K}}$. Therefore, the compactness of the set of all $\{p_k\}_{\in\mathcal{K}}$ satisfying $p_k\geq 0~{\rm and}~\sum_{k\in\mathcal{K}}p_k=1$ indicates the compactness of $\mathcal{W}^*$.
} so that there always exists a large enough upper bound for all values of $\norm{\nabla_{\bs{w}}F_k(\bs{w})}_2$ taken over the set.

We now present a formal bound of the absolute difference of the payment $\mathcal{P}_k^*$ and the exact VCG payment in the following:

\begin{proposition}[Proof in Appendix \ref{AP4}]\label{P4}
 The payment accuracy loss (the absolute difference between $\mathcal{P}_k^*$ and the VCG payment $\mathcal{P}_k^{\rm VCG}$ in \eqref{VCG}) is bounded by:
\begin{align}
&~\left|\mathcal{P}_k^* - \mathcal{P}_k^{\rm VCG}\right| \leq \frac{1-p_k}{p_k}L_g L_f^2(T_2+1)\eta_2^2,~\forall k\in\mathcal{K}.
\end{align}
\end{proposition}
Intuitively, as Proposition \ref{P4} indicates,  $\mathcal{P}_k^*$ converges to the VCG payment in \eqref{VCG} as the step sizes converge to zeros, i.e., $\eta_2\rightarrow 0$.





In the following, we study the iteration complexity of the FFL mechanism, starting with the following lemma:
\begin{lemma}[Proof in Appendix \ref{Proof-L2}]\label{L5}
With equal weights ($p_k=1/K$ for all $k\in\mathcal{K}$), the (Euclidean) distance between $\bs{w}_{-k}^o$ and $\bs{w}^o$ satisfies
\begin{align}
    \norm{\bs{w}_{-k}^o-\bs{w}^o}_2\leq \frac{L_f}{\mu K},~\forall k\in\mathcal{K}.
\end{align}
\end{lemma}
Lemma \ref{L5} implies that the distance of $\bs{w}_{-k}^o$ to $\bs{w}^o$ is  inversely proportional to the total number of agents $K$.
In the following throughout this paper, we choose equal weights $p_k=1/K$ for all $k\in\mathcal{K}$.
Let $\lceil \cdot \rceil$ be the ceiling operator such that $\lceil x \rceil$ is the smallest non-negative integer that is no less than $x$. Define $G\triangleq\norm{\bs{w}[0]-\bs{w}^o}_2$. 
Based on Lemma \ref{L5}, we now have one of the main results of this work:
\begin{theorem}\label{T1}
Set 
$\eta_1={1}/{L_g}$, $\eta_2={1}/(KL_g)$, and $T_1\geq \frac{2\ln(KG/{\Delta})}{\ln((1-\mu/L_g)^{-1})}$. Set
the number of iterations $T_2=\left\lceil\ln\left( \frac{(L_f+\Delta\mu)^2 L_g}{\mu^2K \epsilon}
    \right)/\ln\left(\frac{L_g}{L_g-\mu}\right)\right\rceil$, for any $K>0$ and $\epsilon>0$  such that    $ \left[\ln\left( \frac{(L_f+\Delta\mu)^2 L_g}{\mu^2K \epsilon}
    \right)/\ln\left(\frac{L_g}{L_g-\mu}\right),\frac{L_g\epsilon K}{2 L_f^2}\right]$ is not empty, we 
have a bounded payment accuracy loss, satisfying
\begin{align}
    |\mathcal{P}_k[T_2]-\mathcal{P}_k^{\rm VCG}|\leq \epsilon,~\forall k\in\mathcal{K}.
\end{align}
and a constant time complexity of
\begin{align}
    KT_2\leq \left( 1+\frac{(L_f+\Delta\mu)^2 L_g}{2\mu^2 e\epsilon}
    \right)/\ln\left(\frac{L_g}{L_g-\mu}\right)=\mathcal{O}(1).
\end{align}
\end{theorem}

Note that, for every $\epsilon$, there always exists a large enough $K$ such that the interval
\begin{align}
    \left[\ln\left( \frac{(L_f+\Delta\mu)^2 L_g}{\mu^2K \epsilon}
    \right)/\ln\left(\frac{L_g}{L_g-\mu}\right),\frac{L_g\epsilon K}{2 L_f^2}\right] \label{Empty}
\end{align}
is not empty.

Theorem \ref{T1} implies that, by ensuring the number of iterations in Phase I to satisfy $T_1=\mathcal{O}(\log(K))$,
the time complexity of the payment computation phase in Algorithm \ref{Algo1}  is in fact $K T_2=\mathcal{O}(1)$ with respect to $K$.  In particular, a sufficiently large $K$ such that $K\geq (L_f+\Delta \mu)^2 L_g/(2\mu^2 \epsilon)$ ensures that $ |\mathcal{P}_k[0]-\mathcal{P}_k^{\rm VCG}|\leq \epsilon$ for all  $ k\in\mathcal{K}$. In other words, when $K$ is sufficiently large, we can set $T_2=0$, in which case Phase II of Algorithm \ref{Algo1} will end without any iterations.
Intuitively, the gradient-based nature benefits significantly from the Euclidean distance (between the $\bs{w}_{-k}^o$ and $\bs{w}^o$) inversely proportional to $K$.  On the other hand, since the per-iteration communication complexity is $\mathcal{O}(K)$, the overall communication complexity is also $\mathcal{O}(K)$.

\subsection{Properties}

In this subsection, we will show that the FFL mechanism satisfies (E2) and (E3) approximately and (E4) exactly.
We first introduce the following definition of (probably approximate) faithfulness as to achieve (E2) approximately:
\begin{definition}[Faithfulness]
A mechanism $\mathcal{M}$ is $(\tilde\epsilon,\tilde\delta)$-faithful if the following bound holds for all agents $k\in\mathcal{K}$,
\begin{align}
    {\rm Pr}\left\{\mathbb{E}[J_k(s_k^m,\bs{s}_{-k}^m)]-\min_{A_k\in\mathcal{A}_k}\mathbb{E}[J_k(A_k,\bs{s}_{-k}^m)]\leq \tilde\epsilon\right\}\geq 1-\tilde\delta,
\end{align}
where $J_k(\cdot)$ is the overall loss introduced in Definition \ref{Def1}.
\end{definition}
That is, the $(\tilde\epsilon,\tilde\delta)$-faithfulness suggests that, when all other agents are following the suggested protocol, the incentive for agent $k$ to deviate from doing so is small with a high probability of $1-\tilde\delta$. The bound $\tilde\epsilon$ may depend on $\tilde\delta$ and the number of agents' data samples $\{n_k\}_{k\in\mathcal{K}}$ and is anticipated to vanish as $n\rightarrow \infty$. It follows that:

\begin{proposition}[Faithful Implementation]\label{P5}
If we choose $\eta_1={1/L_g}$, $\eta_2=\frac{1}{KL_g}$, $T_1\geq \frac{2\ln(KG/\Delta)}{\ln((1-\mu/L_g)^{-1})}$, and $T_2=\left\lceil\ln\left( \frac{(L_f+\Delta\mu)^2 L_g}{\mu^2K \epsilon}
    \right)/\ln\left(\frac{L_g}{L_g-\mu}\right)\right\rceil$ for all $\epsilon>0$ such that    $ \left[\ln\left( \frac{(L_f+\Delta\mu)^2 L_g}{\mu^2K \epsilon}
    \right)/\ln\left(\frac{L_g}{L_g-\mu}\right),\frac{L_g\epsilon K}{2 L_f^2}\right]$ is not empty, then the FFL mechanism is $\left(\tilde\epsilon ,\tilde\delta\right)$-faithful, where $\tilde\epsilon$ satisfies
\begin{align}
\tilde\epsilon= 2{\epsilon} &+ K\Phi(\tilde{\delta}),~~\forall \tilde{\delta}\in(0,1),\label{faith}
\end{align}
where $\Phi(\delta)$ is defined in \eqref{Phi}.
\end{proposition}
We note that $K\Phi(\tilde{\delta})\leq C_1\log(1/\delta)\sum_{k\in\mathcal{K}}\frac{ {1}}{n_k}+C_2$ for some constants $C_1$ and $C_2$ (independent of $K$) when $T_1=\mathcal{O}(\log(K))$.
We present the proof of Proposition \ref{P5} in Appendix \ref{AP5}.
The proof of Proposition \ref{P5} is an application of Theorem \ref{T1} and Proposition \ref{P1}.
An interesting observation is that, different from \eqref{Untruthful}, data distributions $\{P_k(\cdot)\}$ do not appear in \eqref{faith}. Thus, we remark that:
\begin{remark}
 The faithful implementation property achieved by the FFL mechanism is robust against non-i.i.d. data. This differs substantially from the classical federated learning settings, in which non-typical agents may have incentives to manipulate federated learning algorithms, as shown in Propositions \ref{P2} and \ref{P3}.
\end{remark}



To show the voluntary participation property, we consider the following probabilistic inequalities. From \eqref{SE-L}, the following inequality holds with a probability of $1-\delta$, for all $k\in\mathcal{K}$,
\begin{align}
    & E_k(\bs{w}^L_k)-\min_{\bs{w}} E_k(\bs{w}) \nonumber\\
    \leq & \left(\sum_{j\neq k}\frac{p_j^2}{n_j(1-p_k)^2}+\frac{p_k}{n_k}\right)\frac{L_{\ell}^2d\log(2Kd/\delta)}{4\mu}
    +\sum_{j\neq k} \frac{p_j}{p_k}\min_{\bs{w}}E_j(\bs{w})\nonumber\\
    &-\min_{\bs{w}}\sum_{j\neq k} \frac{p_j}{p_k} E_j(\bs{w})+2\sum_{j\neq k}p_j\norm{P_j-\bar{P}_{\mathcal{K}\backslash\{k\}}}\triangleq {\rm RB}^{L}_{k,\delta}. \label{RBL}
\end{align}
Following \eqref{Ds}, the following inequality holds with a probability of $1-\delta$, we have that the difference between each agent $k$'s expected overall loss from participation into the FFL mechanism and  $\min_{\bs{w}}E_k(\bs{w})$ satisfies
\begin{align}
    & E_k(\bs{w}^o)+\mathbb{E}[\mathcal{P}_k^{\rm VCG}]-\min_{\bs{w}}E_k(\bs{w}) \nonumber\\
    \lessapprox &  \sum_{j\in\mathcal{K}}\frac{p_j^2}{n_j p_k}\frac{L_{\ell}^2 d\log(2Kd/\delta)}{4\mu}
    +\sum_{j\neq k} \frac{p_j}{p_k}\min_{\bs{w}}E_j(\bs{w})\nonumber\\
    &-\min_{\bs{w}}\sum_{j\neq k} \frac{p_j}{p_k} E_j(\bs{w})+2\sum_{k\in\mathcal{K}}p_k\norm{P_k-\bar{P}_{\mathcal{K}}}\triangleq {\rm RB}^{FFL}_{k,\delta}. \label{RBFFL}
\end{align}

Based on the above probabilistic bounds and Assumption \ref{Assum5}, we can derive the following result:
\begin{proposition}[Risk-Bound-Based Voluntary Participation]\label{P6}
If we choose $\eta_1={1/L_g}$, $\eta_2={1}/(KL_g)$, $T_1\geq \frac{2\ln(KG/\Delta)}{\ln((1-\mu/L_g)^{-1})}$, and $T_2=\left\lceil\ln\left( \frac{(L_f+\Delta\mu)^2 L_g}{\mu^2K \epsilon}
    \right)/\ln\left(\frac{L_g}{L_g-\mu}\right)\right\rceil$ for all $\epsilon>0$ such that    $ \left[\ln\left( \frac{(L_f+\Delta\mu)^2 L_g}{\mu^2K \epsilon}
    \right)/\ln\left(\frac{L_g}{L_g-\mu}\right),\frac{L_g\epsilon K}{2 L_f^2}\right]$ is not empty, then
    the following inequality holds, for all $k\in\mathcal{K}$,
\begin{align}
 {\rm RB}^{FFL}_{k,\delta} \leq~&{\rm RB}^{L}_{k,\delta} +\epsilon\nonumber\\
 &+\frac{2K L_g}{\mu}\left(1-\frac{\mu}{L_g}\right)^{T_1} ( F(\bs{w}[0])-F(\bs{w}^o)).
\end{align}
\end{proposition}
We present the proof of Proposition \ref{P6} in Appendix \ref{AP6}.
Intuitively,
although agents do not know their exact data distributions, Proposition \ref{P6} suggests that the risk bound of the FFL mechanism is smaller than that of local learning plus $\epsilon$. This incentivizes agents to voluntarily participate into the FFL mechanism, achieving (E3).

Finally, we show that the FFL mechanism satisfies (E4) in the following:
\begin{proposition}[Budget Balance, Proof in Appendix \ref{AP7}]\label{P7}
The FFL mechanism $\mathcal{M}$ achieves budget balance (E4):
\begin{align}
    \sum_{k\in\mathcal{K}}\mathcal{P}_k^*\geq 0.
\end{align}
\end{proposition}

To summarize, our FFL algorithm and the FFL mechanism achieve all the desired economic properties of (E1)-(E4). In addition, our FFL mechanism is also scalable as it only incurs an iteration complexity of $\mathcal{O}(\log(K))$ preserves agent privacy as it does not directly require agents to reveal their empirical risks or training data.

\section{Differentially Private Faithful Federated Learning} \label{Sec:DP-FFL}

In this section, we aim to design a faithful federated learning mechanism achieving
a more rigorous guarantee of privacy.
We design a scalable VCG payment, and leverage differential privacy and secure multi-party computation to design a differentially private faithful federated learning algorithm  and the corresponding mechanism.

\subsection{Scalable VCG Payment}

The  VCG payment in \eqref{VCG} for  a  system with $K$ agents requires one to solve $K+1$ optimization problems, which incurs considerable communications and computation overheads for a large-scale system. We showed in Section \ref{Sec:FFL} that the time complexity of solving these problems using a gradient-based algorithm is $\mathcal{O}(\log(K))$ with respect to $K$.
However, to achieve differential privacy, as we will show next, one relies on gradient perturbation under which the number of iterations for each problem no longer decreases in $K$.
To this end, we introduce a scalable approximation of the VCG payment in \eqref{VCG} by reducing the number of problems to be solved.

We formally introduce the Scalable VCG payment in the following:
\begin{definition}[Scalable VCG Payment]\label{Def2}
We randomly divide the set of agents $\mathcal{K}$  into $\mathcal{L}=\{1, 2, ..., L\}$ disjoint clusters. Each cluster is indexed by $l$ and denoted by $\mathcal{C}_l$. We properly divide $\mathcal{K}$ in such a way that each cluster $\mathcal{C}_l$ has either 
$\lceil \frac{K}{L}\rceil$ or $\lfloor \frac{K}{L}\rfloor$ agents.\footnote{As an example, a set of $K=18$ agents can be divided into the following $4$  clusters:
    $\mathcal{C}_1=\{1,4,5,14\}, \mathcal{C}_2=\{2,3,7,10,15\},  \mathcal{C}_3=\{11,13,16,17\},$ and $\mathcal{C}_4=\{6,8,9,12,18\}.$}

The scalable VCG payment for each agent $k$ is
\begin{align}
    \mathcal{P}^{\rm S}_k=\frac{1}{p_k}\sum_{j\neq k}p_j\left( F_j(\bs{w}^o)-F_j(\bs{w}^o_{l})\right),~\forall k\in\mathcal{C}_l,~l\in\mathcal{L},
\end{align}
where 
\begin{align}
     \bs{w}_{l}^o=\arg \min_{\bs{w}}\sum_{k\in \mathcal{K}/\mathcal{C}_l}p_kF_k(\bs{w}),~\forall l\in\mathcal{L}.
\end{align}
\end{definition}

Hence, we approximate $\bs{w}_{-k}^o$ for all $k\in\mathcal{C}_l$ by $\bs{w}^o_l$. In this case, instead of solving $K$ optimization problems, we only need to solve $L$ optimization problems. In the following theorem, we introduce a proper way to select $L$:
\begin{theorem}[Proof in Appendix \ref{AT3}]\label{T3}
If we select 
\begin{align}
    L\geq \min\left\{K,\sqrt{\frac{L_g (K-1)}{2 \epsilon }}\frac{L_f}{\mu}\right\}=\mathcal{O}\left(\sqrt{\frac{K}{\epsilon}}\right), \label{M}
\end{align}
then the Scalable VCG Payment in Definition \ref{Def2} leads to an approximation error of
\begin{align}
   \left|\mathcal{P}_k^{\rm S}-\mathcal{P}_k^{\rm VCG}\right|\leq \epsilon,~\forall k\in\mathcal{K},
\end{align}
where $\mathcal{P}_k^{\rm VCG}$ is the VCG payment for agent $k$ in \eqref{VCG}.
\end{theorem}
The proof of Theorem \ref{T3} uses a similar technique to that of Lemma \ref{L5}, as  $\norm{\bs{w}^o_{-k}-\bs{w}^o_{l}}_2$ is inversely proportional to the number of agents within each cluster $\mathcal{C}_l$.

Theorem \ref{T3} indicates that, to maintain a bounded approximation error, the  number of optimization problems to be solved grows at a square root rate, compared to the classical VCG mechanism with a linear rate. Therefore, the Scalable VCG payment in Definition \ref{Def2} allows us to design a more scalable mechanism when we cannot rely on a gradient-based algorithm.

\subsection{Differentially Private FFL Algorithm and Mechanism}

\begin{algorithm}[t]
\caption{Differentially Private Faithful Federated Learning (DP-FFL)}
\label{Algo2}
\SetAlgoLined
The server initializes $\bs{w}[0]$ and step sizes $\eta_1, \eta_2$\;

\tcp{Phase I: Federated learning phase}\label{l1}

 \For{\textbf{iterations} $t\in\{0, 1,..., T_1\}$}{ 
The server broadcasts $\bs{w}[t]$ to all agents\;

Each agent $k$ computes and reports its gradient to the server $\nabla {F}_k(\bs{w}[t]))$\;

The server \textit{securely aggregates agents' gradients, adds noise}, and updates the model according to:
        \begin{align}
 \bs{w}[t+1]=\bs{w}[t]- \eta_1\left(\sum_{k\in\mathcal{K}} p_k\nabla {F}_k(\bs{w}[t])+\bs{n}\right), \label{model}
        \end{align}
where $\bs{n}$ is a random vector sampled from $\mathcal{N}(\bs{0},\sigma^2 \bs{I}_d)$ and $\sigma^2$ is given  in \eqref{noise}\label{noise1}\;
 }
 \textit{\textbf{Return}} $\bs{w}^*=\bs{w}[{T}_1]$\; \label{l7}
 
 \For{\textbf{clusters} $l\in\mathcal{L}$}{
  \tcp{Phase II: Payment computation phase}\label{l8}
  
    The server initializes the model $\bs{w}_{l}[0]=\bs{w}^*$\;
   
 \For{\textbf{iterations} $t\in\{0, 1,..., T_2\}$}{ 

 The server broadcasts $\bs{w}_l[t]$ to all agents not in $\mathcal{C}_l$\;
 
Each agent $k\notin \mathcal{C}_l$ computes and reports its gradient to the server $\nabla {F}_k(\bs{w}_l[t])$\;\label{l12}

 The server \textit{securely aggregates agents' gradients and adds noise} to update the model:
        \begin{align}
      \bs{w}_{l}[t+1]&=\bs{w}_{l}[t]- \eta_2\left(\sum_{j\in \mathcal{C}_l} p_j\nabla {F}_j(\bs{w}_{l}[t])+\bs{n}\right),
        \end{align}
        where $\bs{n}$ is a random vector sampled from $\mathcal{N}(\bs{0},\sigma^2 \bs{I}_d)$ and $\sigma^2$ is given  in \eqref{noise}\label{noise2}\;
 }
 
Each agent $k\in\mathcal{K}$ reports the value $F_k(\bs{w}^*)-F_k(\bs{w}_{l}[T_2])$ and the server computes
\begin{align}
\mathcal{P}_k^*=\sum_{j\neq k} \frac{p_j}{p_k}\left({F}_j(\bs{w}^*)- {F}_j(\bs{w}_{l}[T_2])\right)+n_{P,k},\label{payment-Algo2}
 \end{align}
 for all $k\in\mathcal{C}_l$,
 where $n_{P,k}$ is sampled from $\mathcal{N}(0,\sigma_P^2)$ and $\sigma_P^2$ is given  in \eqref{noise}\label{noise3}\;
 
 {\textbf{Return}} the payment $\mathcal{P}_k^*$ for agent $k\in\mathcal{C}_l$\; 
 }
 \label{l15}
\end{algorithm}

Motivated by a recent differentially private federated learning algorithm in \cite{SDP}, we next introduce the techniques to ensure differential privacy by combining secure multi-party computation (to aggregate agents' local gradients) and gradient perturbation. Jayaraman \textit{et al.} in \cite{SDP} showed that this allows the server to add only a single noise copy, and can outperform the algorithms requiring local gradient perturbation before aggregation.


Before we describe and analyze the algorithm, we first formally introduce the following concepts:
\subsubsection{Differential Privacy} 

We aim to guarantee \textit{differential privacy}, which is a crypographically-motivated notion of privacy \cite{DP}. We define $\alpha\geq 0$ as privacy risk. Formally, we have:\footnote{Note that $(\alpha,\beta)$-differential privacy is also known as $(\epsilon,\delta)$-differential privacy, as in \cite{SDP}.}
\begin{definition}[$(\alpha,\beta)$-differential privacy (DP)]
Let $\alpha$ be a positive real number and 
${\mathcal {Z}}$ be a randomized algorithm. The algorithm 
${\mathcal {Z}}$ is said to provide $(\alpha,\beta)$-DP if, for any two datasets $\mathcal{D}_1$ and $\mathcal{D}_2$
that differ on a single element, ${\mathcal {Z}}$ satisfies
\begin{align}
\displaystyle \Pr[{\mathcal {Z}}(\mathcal{D}_1)= y]\leq \exp \left(\alpha \right)\cdot \Pr[{\mathcal {Z}}(\mathcal{D}_2)= y]+\beta,~\forall y. 
\end{align}
\end{definition}
The above definition is reduced to the $\alpha$-DP when $\beta=0$, as in \cite{DPSGD}. As we will show, one can achieve $(\alpha,\beta)$-DP by adding noise sampled from Gaussian distributions to gradients.

\subsubsection{Secure Multi-Party Computation} 
For preserving the privacy of agents' inputs without revealing them to others, the server aims to securely aggregate their gradients in each iteration.\footnote{Note that MPC protocols are only able to protect the training data during the learning process, whereas the resulting model of a federated learning algorithm still relies on the differential privacy techniques (by adding noise in our proposed DP-FFL algorithm) against inferring private data of each agent.}
To this end, we consider secure multi-party computation (MPC) protocols that enable one to jointly aggregate their private inputs. Examples of these are protocols that employ cryptographic techniques (e.g., homomorphic encryption and  secret sharing).
In this work, we do not focus on improving or evaluating the MPC protocols, since the methods we propose can be implemented using standard MPC techniques.\footnote{A concrete example of the standard MPC technique in federated learning frameworks can be found in \cite{SDP,Privacy}.}

\subsubsection{Algorithm and Mechanism Description} 
We are ready to introduce the \textbf{Differentially Private Faithful Federated Learning (DP-FFL) algorithm} in Algorithm \ref{Algo2}, which also consists of two phases.
Compared to the FFL algorithm in Algorithm \ref{Algo1}, we add noise  in lines \ref{noise1}, \ref{noise2}, and \ref{noise3} based on
\begin{align}
  \sigma^2&=\frac{16L_f^2 (T_1+KT_2)\log(1/\beta)}{K^2 n_{(1)}^2\alpha^2}\\
  \sigma_P^2&=\frac{16 K \log(1/\beta)}{ n_{(1)}^2\alpha^2},\label{noise}
\end{align}
where $n_{(1)}$ is the size of smallest local dataset among all agents $\mathcal{K}$.
We also adopt the Scalable VCG payment from Definition \ref{Def2} in the payment computation phase of Algorithm \ref{Algo2}.

In the following, we introduce the  DP-FFL mechanism associated to Algorithm \ref{Algo2}:
\begin{definition}[The DP-FFL Mechanism]
The DP-FFL mechanism $\mathcal{M}^{\rm P}=(\mathcal{A}, \mathcal{S}^m,\bs{w}^*, \mathcal{P})$ satisfies, for all agents $k\in\mathcal{K},$
\begin{align}
    s^m_k=\{\nabla F_k(\bs{w}[t])\}_{t\in \mathcal{T}_{-k}}\quad{\rm and}\quad  A_k=\{\nabla \tilde{F}_k(\bs{w}[t])\}_{t\in \mathcal{T}_{-k}},
\end{align}
where $\mathcal{T}_{-l}=\bigcup_{j\in\{0\}\cup\mathcal{L}\backslash\{l\}}\{1\leq t\leq T_2\}$. The output global model $\bs{w}^*$ and each agent $k$'s payment $\mathcal{P}_k$ are determined in \eqref{model} and \eqref{payment-Algo2}, respectively.
\end{definition}

 We re-define the gradient bound specific for Algorithm \ref{Algo2} in the following:
\begin{definition}[Gradient Bound]\label{Assum4}
The gradient bound $L_f$ is defined as for each agent $k\in\mathcal{K}$,\footnote{For readability, we are overloading the notation $L_f$  to avoid introducing additional parameter names.}
    \begin{align}
  \norm{\nabla_{\bs{w}}F_k(\bs{w})}_2\leq L_f, 
  \end{align}
  for all $\bs{w}\in\{\bs{w}[t]\}_{t\in\{0,1,...,T_1\}}\cup\{\bs{w}_{l}[t]\}_{l\in\mathcal{L}, t\in\{0,1,...,T_2\}}\cup \mathcal{W}^*,$
  where $\{\bs{w}[t]\}_{t\in\{0,1,...,T_1\}}\cup\{\bs{w}_{l}[t]\}_{l\in\mathcal{L}, t\in\{0,1,...,T_2\}}$ are described in Algorithm \ref{Algo2} and $\mathcal{W}^*$ is defined in \eqref{BigW}.
\end{definition}

We now show that DF-FFL achieves the following privacy property:
\begin{proposition}[Proof in Appendix \ref{AP8}]\label{P8}
 Algorithm \ref{Algo2} is $(\alpha,\beta)$-differentially private.
\end{proposition}

Proof of Proposition  \ref{P8} is based on \cite{SDP} and \cite{concen}. 
Similar to Proposition \ref{P3.5}, we can also prove  (E1) for the  DF-FFL mechanism in the following:
\begin{proposition}[Proof in Appendix \ref{AP7.5}]\label{P7.5}
 In Algorithm \ref{Algo2}, if we choose step sizes such that $\eta_1=\eta_2\leq 1/L_g$ and numbers of iterations $T_1=\mathcal{O}\left(\log\left(\frac{ K^2n_{(1)}^2 \alpha^2}{d L_{f}^2 \log(1/\beta)}\right)\right)$ and $T_2=\mathcal{O}\left(\log\left(\frac{ K^2n_{(1)}^2 \alpha^2}{d L_{f}^2 \log(1/\beta)}\right)\right)$,\footnote{the big-$\mathcal{O}$ notation hides other $\log$, $L_g$, and $\mu$ terms.}
 the following \textit{risk bound} is true with a probability of $1-\delta$:
\begin{align}
E(\bs{w}^*)-\min_{\bs{w}} E(\bs{w})\leq & ~C_1\frac{L_{f}^2 d \log (Kn_{(1)})\log(1/\beta)}{K n_{(1)}^2\alpha^2}\nonumber\\
&~+\sum_{k\in\mathcal{K}}\frac{1}{n_k}\frac{L_{\ell}^2d\log(2d/\delta)}{2\mu K^2},
\end{align}
for some constants $C_1>0$.
 \end{proposition}


\subsection{Three-Way Tradeoffs between Privacy, Accuracy, and the Iterations Needed}

   	\begin{figure}[t]
  	\centering
		\subfigure[]{\includegraphics[scale=.22]{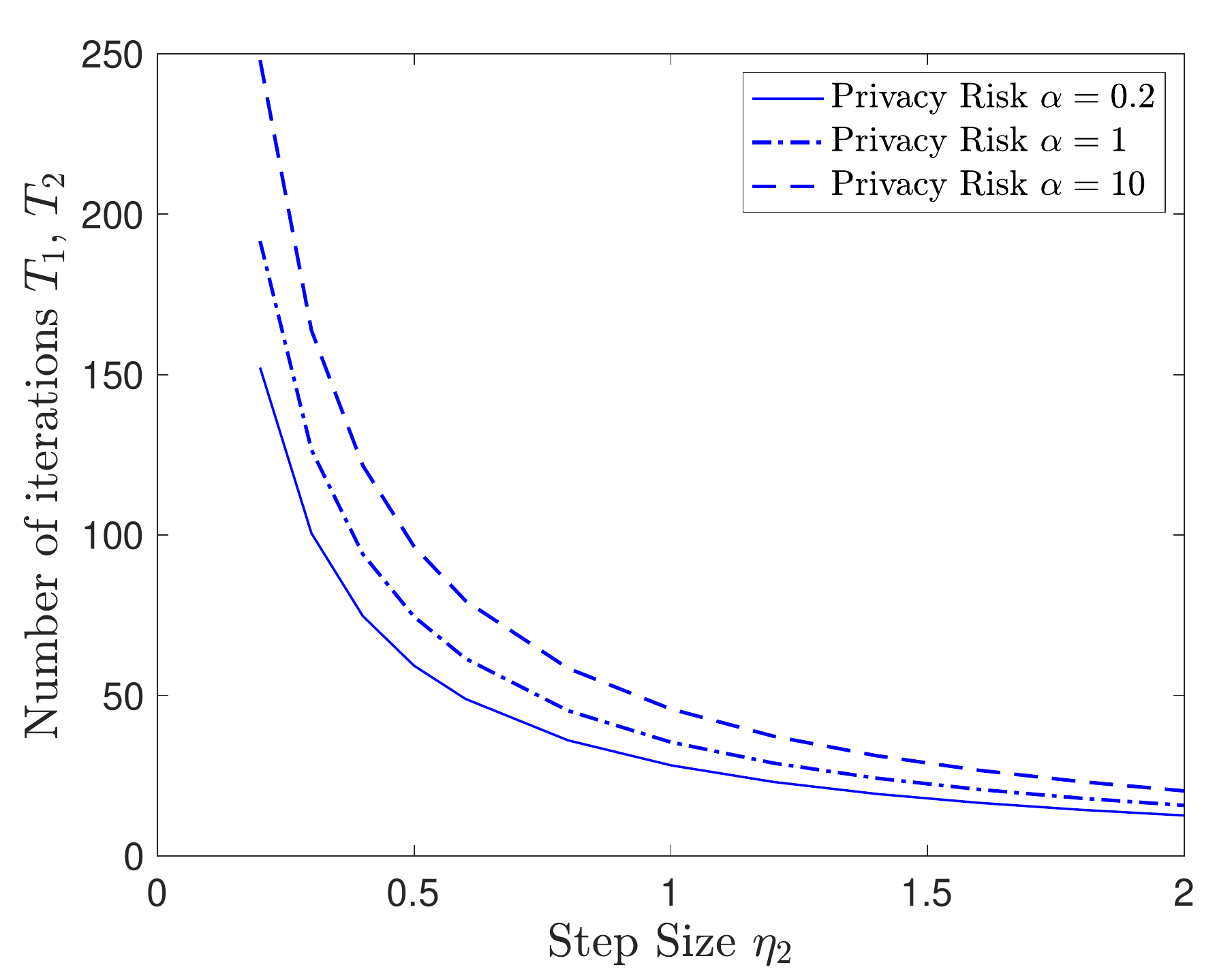}}
			\hspace{0.05\linewidth}
		\subfigure[]{\includegraphics[scale=.22]{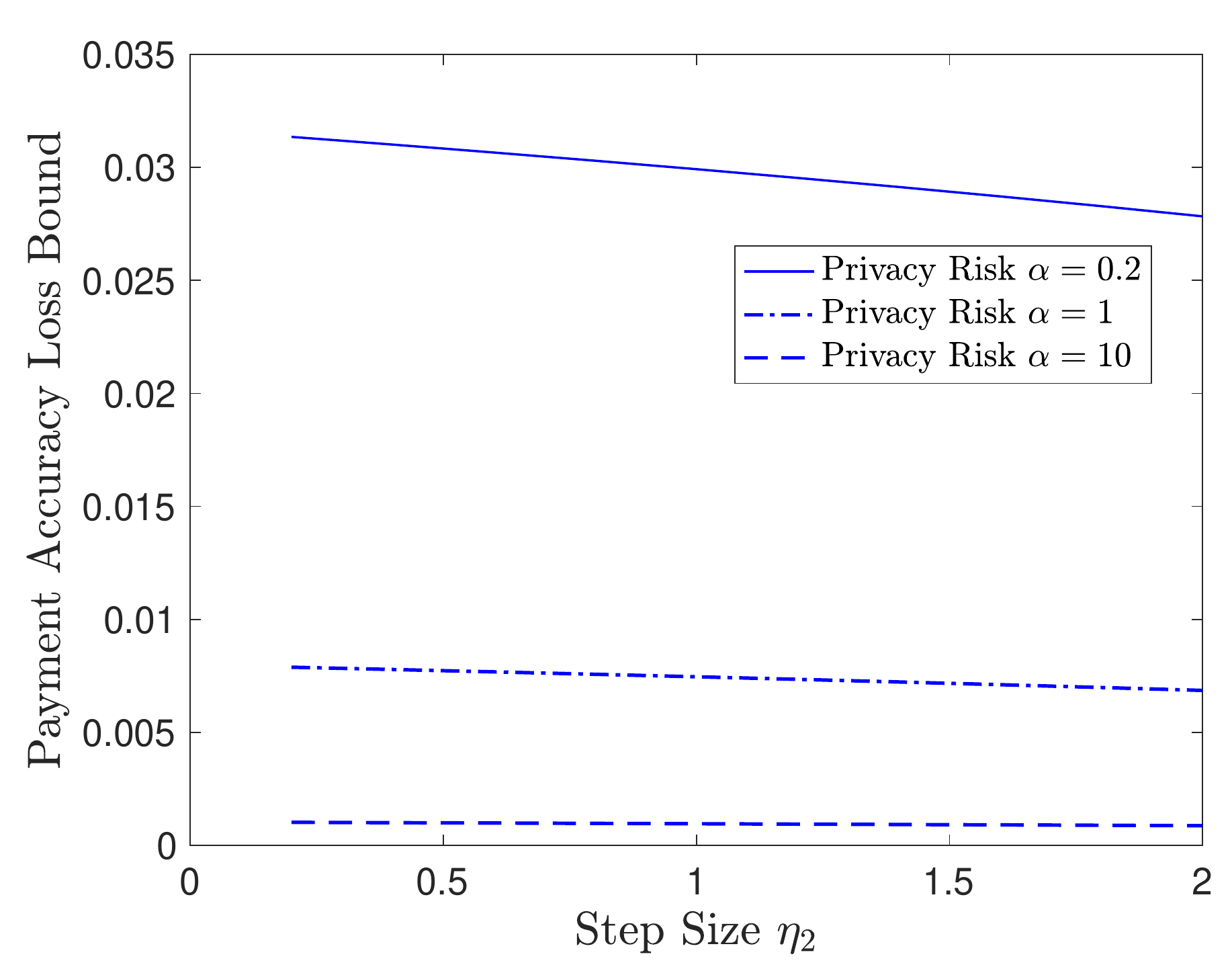}}
		\caption{Impacts of $\eta_2$ and $\alpha$ on the iterations needed $T_1, T_2$ and payment accuracy loss bounds $ \left|\mathbb{E}[\mathcal{P}_k^*]-\mathcal{P}^{\rm VCG}_k\right|$. }\label{Three-Way}
 		\vspace{-0.3cm}
	\end{figure}

We will  discuss three-way performance tradeoffs in the following:
\begin{proposition}[Proof in Appendix \ref{AP9}]\label{P9}
If we choose constant step sizes $\eta_1\leq 1/L_g$, $\eta_2\leq \frac{1}{(K-1)L_g}$, $L\geq \min\left\{K,\sqrt{\frac{L_g (K-1)}{2 \epsilon }}\frac{L_f}{\mu}\right\}$, for any $\epsilon>0$, and the iterations as
\begin{align}
   T_1= T_2=\left\lceil\frac{\log\left(\frac{B\log(C^{-1})}{A}\right)}{\log(C^{-1})}\right\rceil, \label{Convergence11}
\end{align}
then Algorithm \ref{Algo2} leads to a bounded expected payment accuracy loss:
\begin{align}
    \left|\mathbb{E}[\mathcal{P}_k^*]-\mathcal{P}^{\rm VCG}_k\right|\leq &~ A\left(\frac{1+\log(\frac{B\log(C^{-1})}{A})}{\log(C^{-1})}\right)+\epsilon, \label{ZXC}
\end{align}
for all agents $k\in\mathcal {K}$, where 
\begin{subequations}\label{Eq-ABC}
\begin{align}
    A&=\frac{8 \eta_2 d L_f^2 (K+1) \log(1/\beta)}{\mu K^2 (K-1) n_{(1)}^2 \alpha^2},\\
    B&= \frac{(K-1) L_f^2}{2\mu},\\
    C&=1-(K-1)\mu \eta_2.
\end{align}
\end{subequations}

\end{proposition}

Proposition \ref{P9} suggests a three-way tradeoff among privacy $(\alpha, \beta)$, the iterations needed ($T_1$ and $T_2$), and accuracy $ \left|\mathbb{E}[\mathcal{P}_k^*]-\mathcal{P}^{\rm VCG}_k\right|$.
That is, by fixing $\beta$ and properly selecting $\alpha$ and $\eta_2$, one can improve two performance metrics by trading off the third. To see this, we consider the following three scenarios as guidelines to make such a three-way tradeoff:
\begin{enumerate}
    \item \textit{To improve both privacy and iteration complexity by sacrificing accuracy}, one can set $\alpha\rightarrow 0$ to ensures privacy and  set $\eta_2$ as a positive constant. In this case, we have that $A\rightarrow \infty$, and hence 
$T_1=T_2\rightarrow 0$. However, the payment accuracy loss in  \eqref{ZXC} diverges to infinity, i.e., $\left|\mathbb{E}[\mathcal{P}_k^*]-\mathcal{P}^{\rm VCG}_k\right|\rightarrow \infty$.
    \item \textit{To improve both accuracy and iteration complexity by sacrificing privacy}, one can set $\alpha$ and $\eta_2$ in  such a way that $\eta_2/\alpha^2$ is constant (and hence $A$ is also constant). Further, by increasing $\eta_2$ (and hence increasing $\alpha$ as well), we have that $C$ decreases, and hence $\log(\log(C^{-1}))/\log(C^{-1})$ decreases. Therefore, such a strategy decreases $T_1$, $T_2$, and $\left|\mathbb{E}[\mathcal{P}_k^*]-\mathcal{P}^{\rm VCG}_k\right|$ at the same time. However, $\eta_2$ is upper-bounded by the maximal step size $\frac{1}{(K-1)L_g}$.
    \item \textit{To achieve perfect accuracy $ \left|\mathbb{E}[\mathcal{P}_k^*]-\mathcal{P}^{\rm VCG}_k\right|\rightarrow 0$, one needs to sacrifice both iteration complexity, privacy, and scalability.} Specifically, by setting $\alpha\rightarrow \infty$, it follows that $A\rightarrow 0$ and hence $T_1=T_2\rightarrow \infty$. Letting $L=K$ as well, we have $ \left|\mathbb{E}[\mathcal{P}_k^*]-\mathcal{P}^{\rm VCG}_k\right|\rightarrow 0$.
\end{enumerate}

{We present a numerical example of Proposition \ref{P9} in Fig. \ref{Three-Way}, which compares the iterations needed and payment accuracy loss bounds at different $\alpha$ and $\eta_2$. An interesting observation is that
the payment accuracy loss bound hardly changes when $\eta_2$ decreases. This results from the fact that $\log((1-(K-1)\mu \eta_2)^{-1})\approx (K-1) \mu \eta_2$, which makes $A/\log(C^{-1})$ almost constant if we only tune $\eta_2$. On the other hand, fixing the step size $\eta_2$, $T_1$ and $T_2$ decrease in $\alpha$ and the payment accuracy loss bound increases in $\alpha$.}


\subsection{Properties}

Collectively, in this subsection, we will show that the DP-FFL algorithm (and the DP-FFL mechanism) satisfies (E2) and (E3) approximately and (E4) exactly. 

\begin{corollary}[Faithful Implementation]\label{P10}
If we choose $\eta_1\leq {1}/{L_g}$, $\eta_2\leq \frac{1}{(K-1)L_g}$ $T_1=T_2=\left\lceil\log\left(\frac{B\log(C^{-1})}{A}\right)/\log(C^{-1})\right\rceil$, and $L\geq \min\left\{K,\sqrt{\frac{L_g (K-1)}{2 \epsilon }}\frac{L_f}{\mu}\right\}$, then the DP-FFL  mechanism is $(\tilde{\epsilon},\tilde\delta)$-faithful, where
    \begin{align}
        \tilde{\epsilon}=&2\epsilon+2A\left(\frac{1+\log(\frac{B\log(C^{-1})}{A})}{\log(C^{-1})}\right)+ \sum_{k\in\mathcal{K}}\frac{1}{n_k}\frac{L_{\ell}^2d\log(2d/\tilde\delta)}{2K\mu}\nonumber\\
        &
        +D\frac{L_f^2 d \log(Kn_{(1)})\log(1/\beta)}{ n_{(1)}^2\alpha^2},\label{Eq-P9}
    \end{align}
and $A, B,$ and $C$ are defined in \eqref{Eq-ABC}, and $D$ is some positive constant.
\end{corollary}

 Corollary \ref{P10} is a direct application of Propositions \ref{P5} and \ref{P9} and Theorem \ref{T3}; the three terms on the right-hand side of \eqref{Eq-P9} come from Theorem \ref{T3}, Proposition \ref{P9}, and Proposition \ref{P5}, respectively.


\begin{corollary}[Risk-Bound-Based Voluntary Participation]\label{C2}
   If we choose $\eta_1\leq \frac{1}{L_g}$, $\eta_2\leq \frac{1}{(K-1)L_g}$ $T_1=T_2=\left\lceil\log\left(\frac{B\log(C^{-1})}{A}\right)/\log(C^{-1})\right\rceil$, and $L\geq \min\left\{K,\sqrt{\frac{L_g (K-1)}{2 \epsilon }}\frac{L_f}{\mu}\right\}$, then the DP-FFL algorithm and the DP-FFL mechanism lead to the following inequality:
\begin{align}
& {\rm RB}^{FFL}_{k,\delta} \leq ~~{\rm RB}^{L}_{k,\delta} +\epsilon +D\frac{L_f^2 d \log(Kn_{(1)})\log(1/\beta)}{ n_{(1)}^2\alpha^2} \nonumber\\&
 + A\left(\frac{1+\log(\frac{B\log(C^{-1})}{A})}{\log(C^{-1})}\right)
\end{align}
for all agents $k\in\mathcal{K},$ where ${\rm RB}^{L}_{k,\delta}$  and ${\rm RB}^{FFL}_{k,\delta}$ are defined in \eqref{RBL} and \eqref{RBFFL}, respectively. 
\end{corollary}

We note that, as implied in Corollary \ref{C2}, performing local learning does not incur privacy loss for individual agents. Therefore, comparing the results in Corollary \ref{C2} and Proposition \ref{P6}, the DP-FFL algorithm leads to a worse risk bound than the FFL algorithm.

Finally, Proposition \ref{P9} also implies that the DP-FFL mechanism achieves budget balance (E4) approximately. We present the detailed analysis in \cite{technical}.

\section{Evaluation} \label{Sec:EVA}

	

  

	



\begin{figure*}[t]

	\begin{minipage}[t]{0.5\linewidth}
		\centering
		\subfigure[]{\includegraphics[scale=.23]{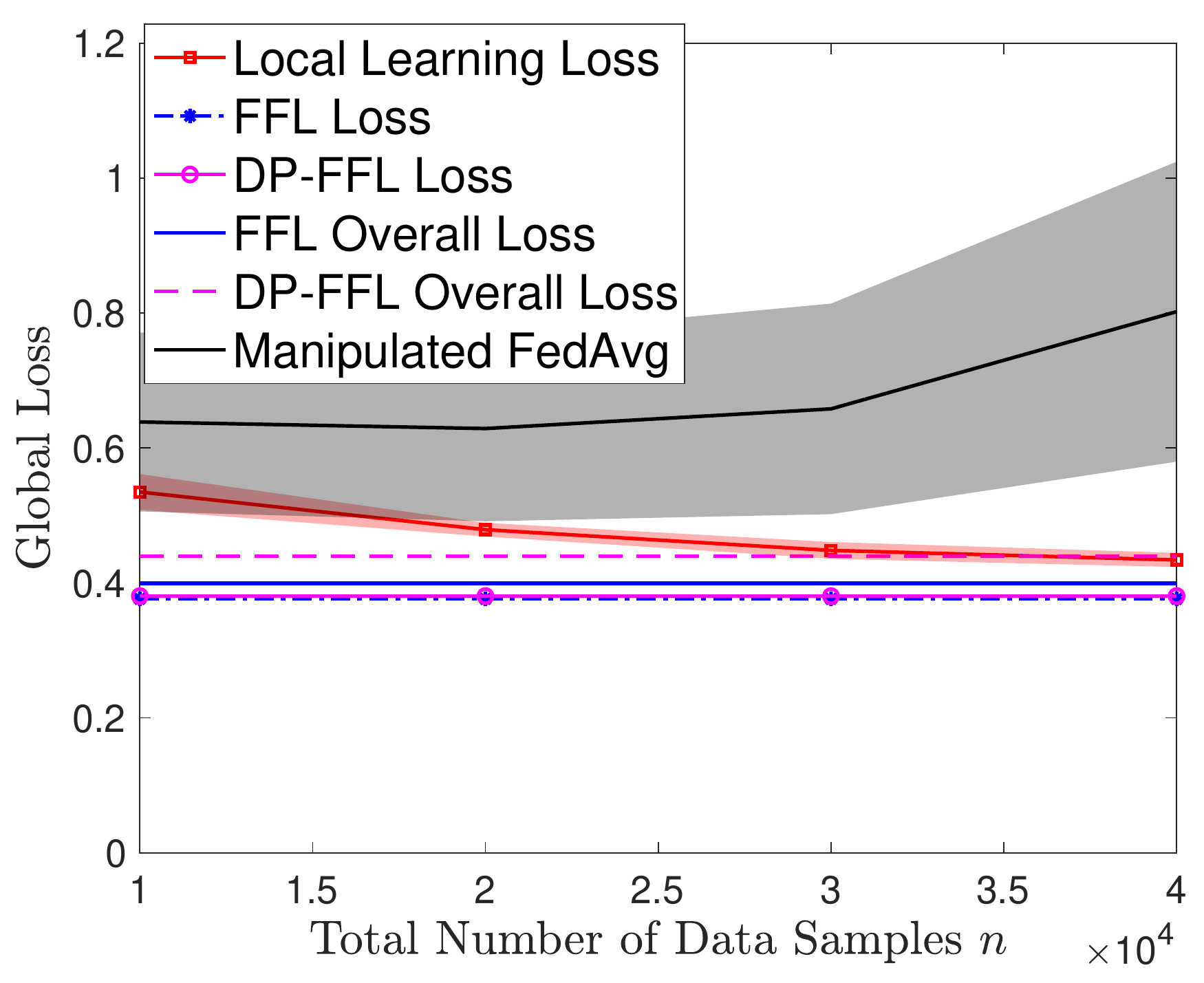}}
		\subfigure[]{\includegraphics[scale=.23]{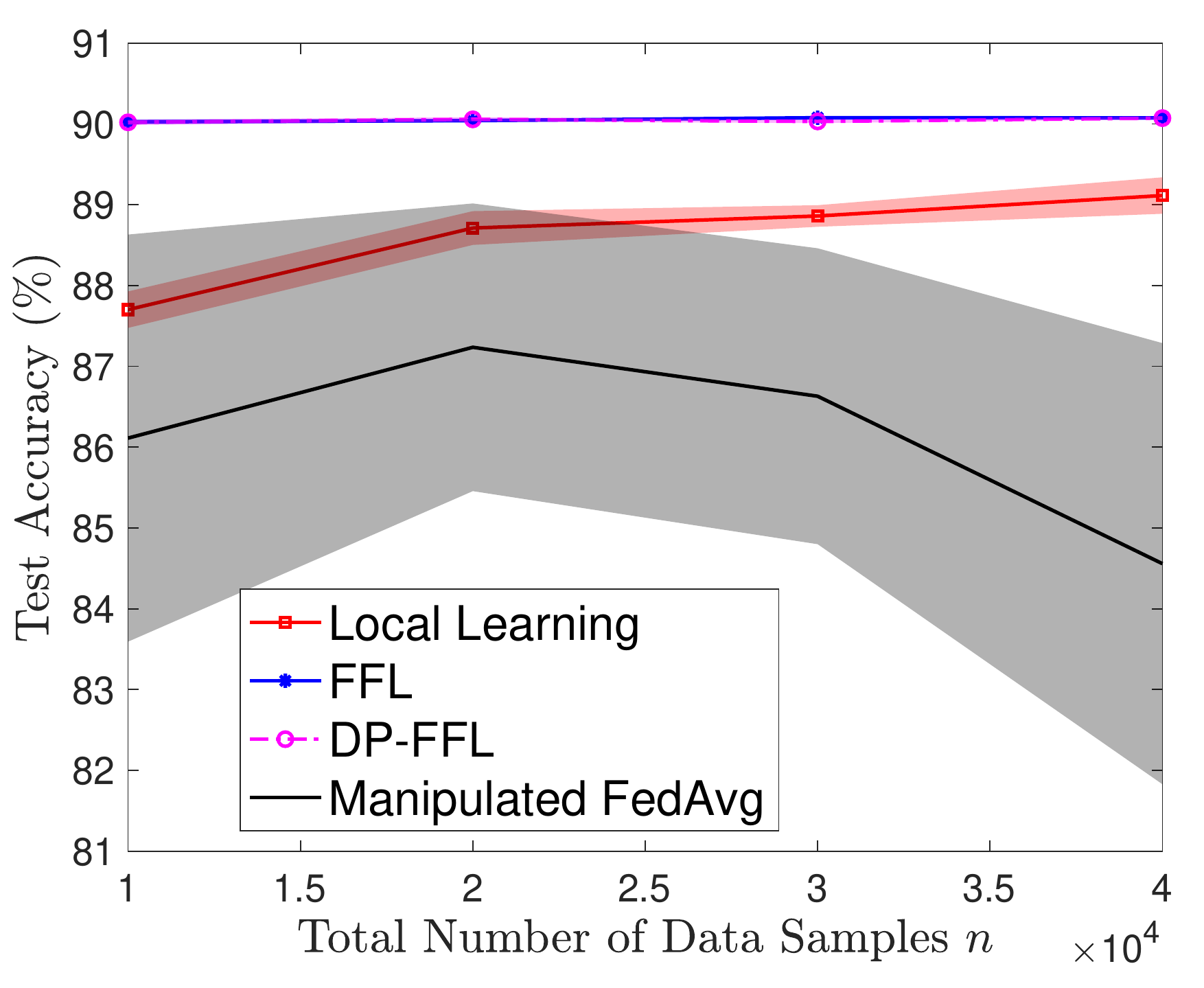}}
		\vspace{-0.1cm}
		\caption{Impacts of training data samples $n$ on (a)
training loss and overall costs and (b) test accuracy. We set $K=10$, $\alpha=0.1$, $\beta=0.01$, $\Delta=0.05$, $T_1=80$, and $T_2=20$. }\label{Fig1}
	\end{minipage}
 	\hspace{0.01\linewidth}
 		\begin{minipage}[t]{0.5\linewidth}
 		  	\centering
		\subfigure[]{\includegraphics[scale=.23]{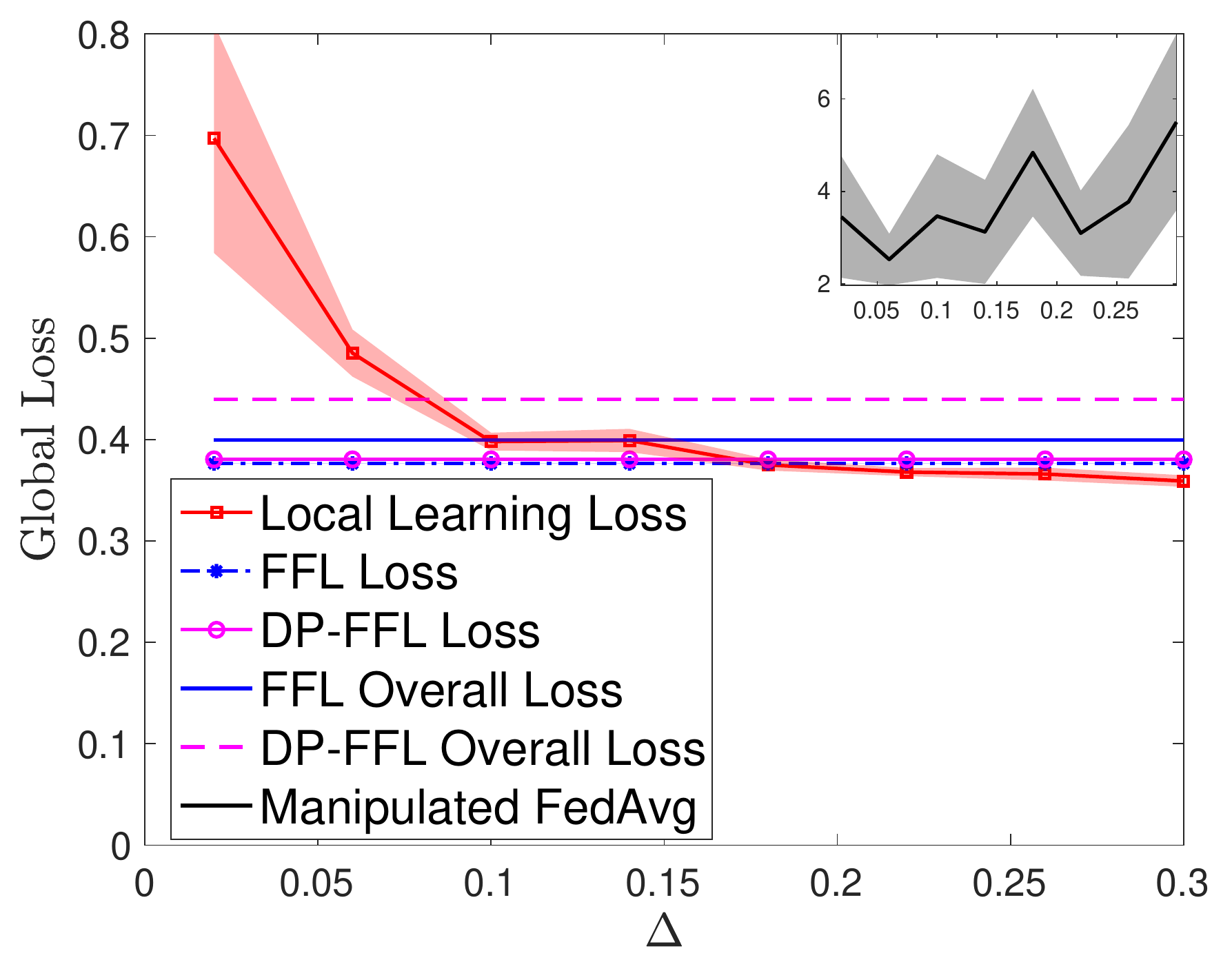}}
		\subfigure[]{\includegraphics[scale=.23]{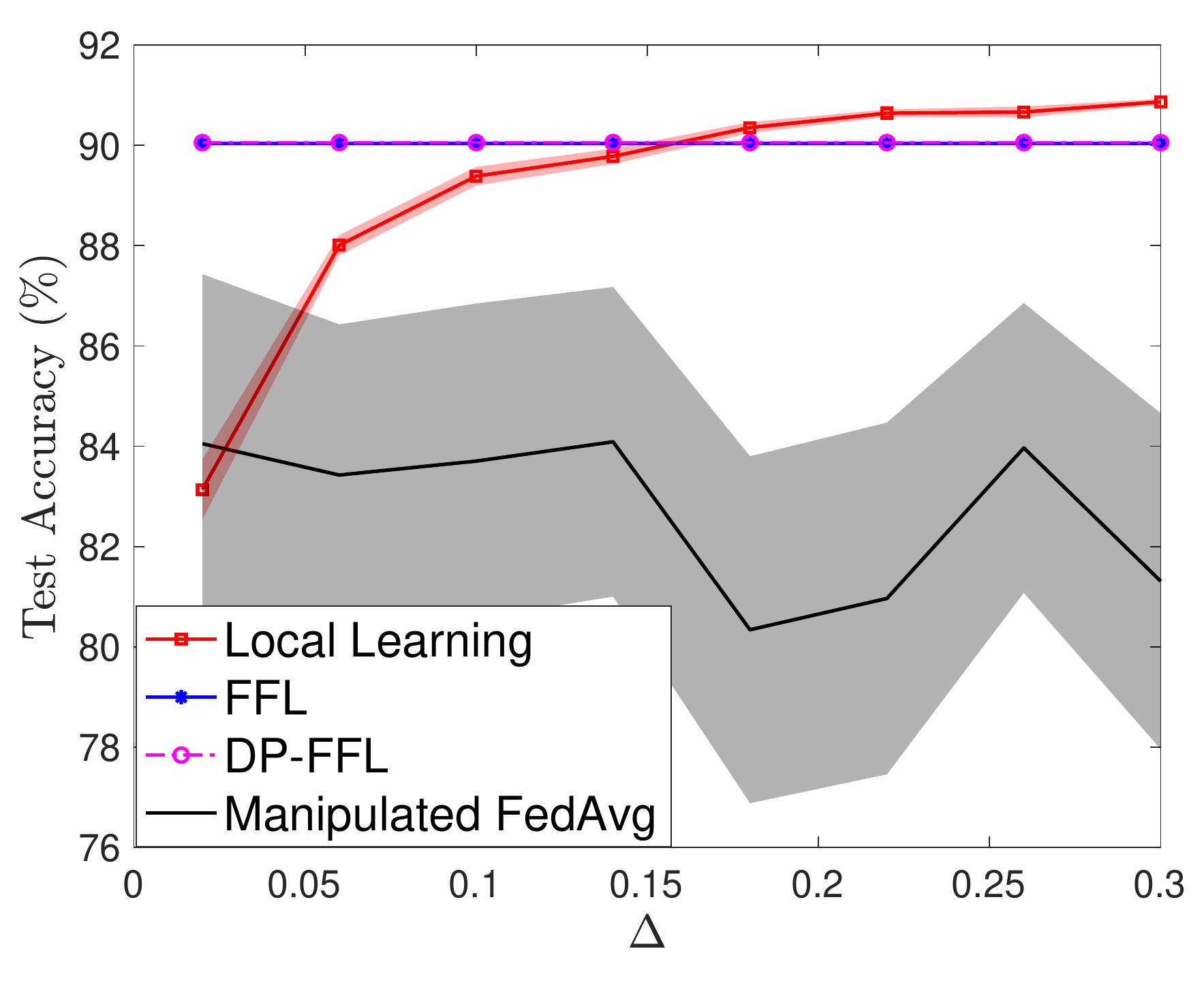}}
		\vspace{-0.1cm}
		\caption{
		Impacts of non-i.i.d. data, characterized by $\Delta$, on (a)
training loss and overall costs and (b) test accuracy. We set $K=10$, $\alpha=0.1$, $\beta=0.01$, $n=10000$, $T_1=80$, and $T_2=20$.}
\label{Fig3}
	\end{minipage}
 	\vspace{-0.3cm}

\end{figure*}

In this section, we evaluate our proposed FFL and DP-FFL mechanisms with $K=10$ agents. We consider regularized multinomial logistic regression for the MNIST dataset with 60,000  training samples and 10,000 testing samples \cite{MNIST}.
We
uniformly randomly allocate samples with label $y$ to all agents whose last digits of their indices are $y$. For each sample allocated to agents, with a probability of $1-\Delta$, we reallocate this sample to a random agent with equal probabilities.
Therefore, the degree of heterogeneity in this non-i.i.d. data can be characterized by $\Delta$; a larger $\Delta$ leads to greater data heterogeneity.

For performance comparison, we compare our proposed FFL and DP-FFL schemes against two benchmarks: i) a gradient-based local learning benchmark, in which agents independently solve \eqref{Eq-P1}, and ii) a manipulated FedAvg benchmark, in which the server intends to execute FedAvg \cite{FL1}, while one agent manipulates the federated learning algorithm by multiplying its gradient report by an amplifying coefficient $\gamma$ in each iteration. We compare the global loss, $F(\bs{w})$, and the weighted average test accuracy achieved by different schemes.

\textbf{Impact of the total number of training samples}:
We study the impact of the number of total training data samples $n$ in Fig. \ref{Fig1}. First, we show that our proposed schemes significantly outperform
the manipulated FedAvg, implying that federated learning manipulated even only by one agent can lead to significant performance loss. 
Therefore, it demonstrates the importance of faithful implementation of federated learning algorithms. Second, both  proposed schemes outperform local learning with respect to either global loss and test accuracy. This also indicates that agents are willing to voluntarily participate in federated learning, as both proposed schemes achieve smaller overall losses.
We observe that the total number of training data samples $n$ has a greater impact on both local learning and the manipulated FedAvg benchmarks than the proposed (FFL and DP-FFL) mechanisms. 
This is because the performances both benchmarks are more sensitive to the sizes of local datasets, compared to federated learning.




\begin{figure}[t]


		\centering
		\subfigure[]{\includegraphics[scale=.23]{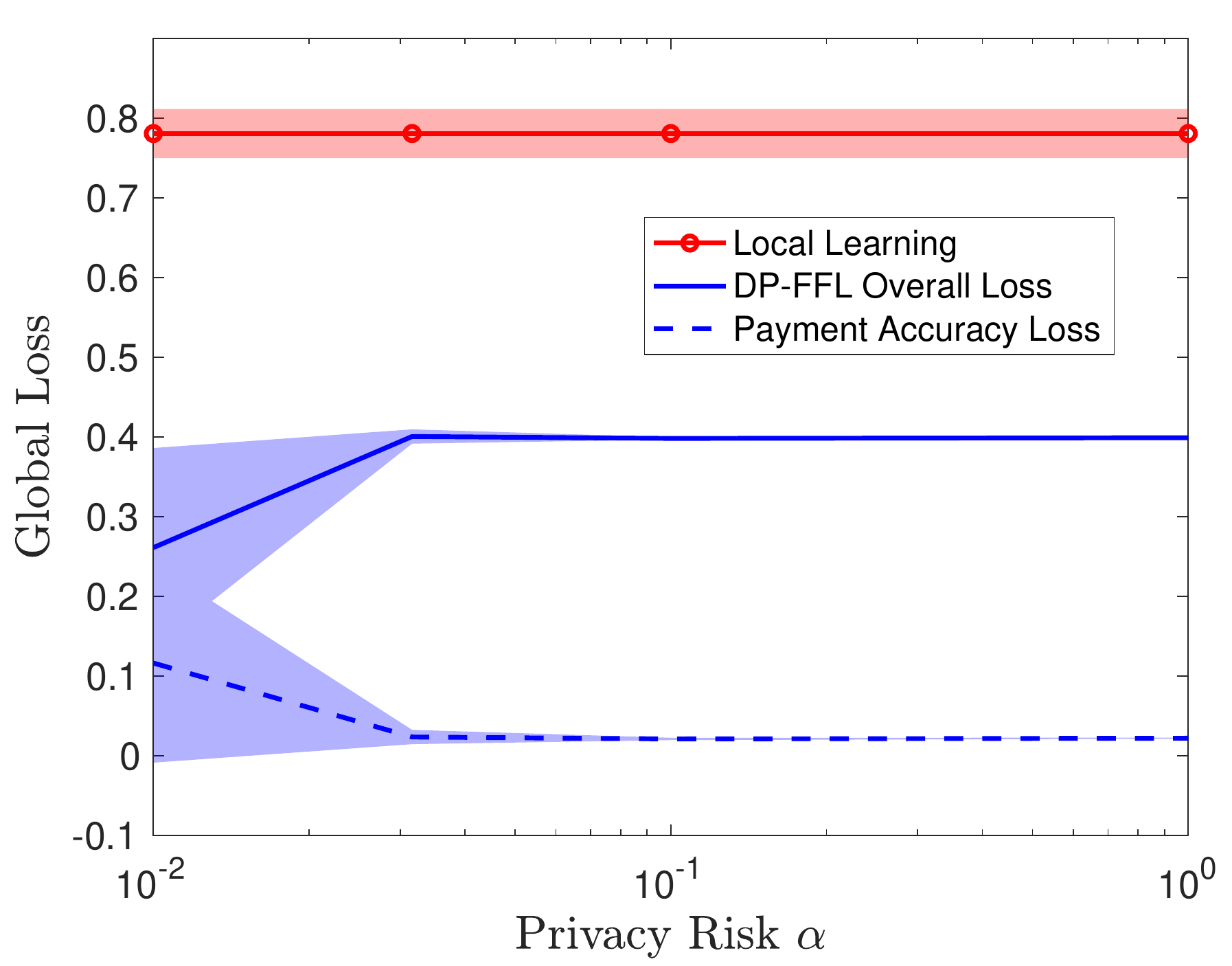}}
		\subfigure[]{\includegraphics[scale=.23]{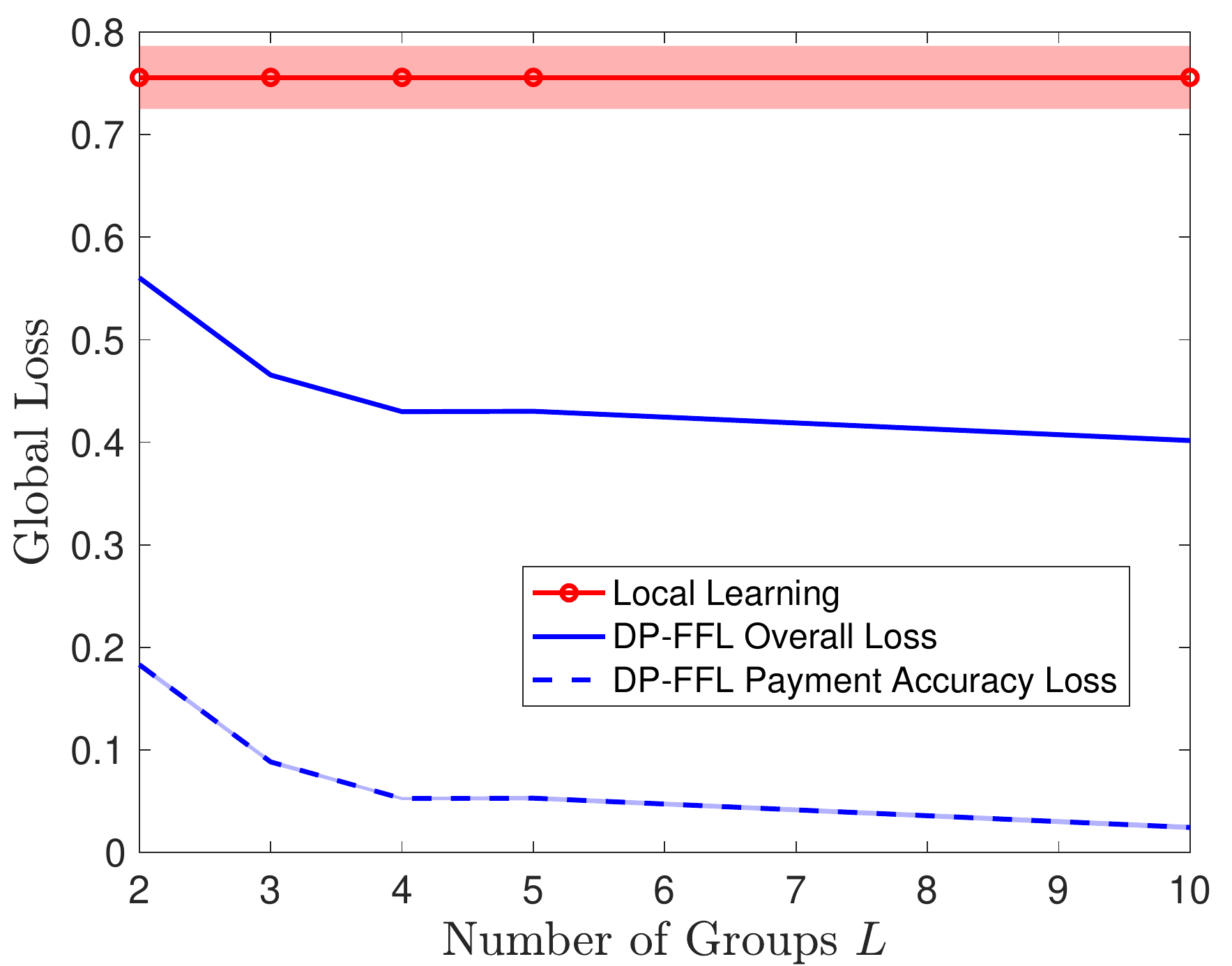}}
 		\vspace{-0.3cm}
		\caption{Impacts of (a) $\alpha$ and (b) $L$. We set $K=10$, $n=10000$, $\beta=0.01$, $\Delta=0.05$, $T_1=80$, and $T_2=20$. We further set $L=10$  in (a) and $\alpha=0.1$ in (b).}\label{Fig5}
\vspace{-0.4cm}
\end{figure}

\textbf{Impact of non-i.i.d. data}:
We next study the impact of varying the heterogeneity in the data given by $\Delta$ 
in Fig. \ref{Fig3}. 
As shown in Fig. \ref{Fig3}, $\Delta$ only has an impact on local learning but not on the proposed mechanisms. 
 In Fig. \ref{Fig3}(a), we show that, in terms of each individual agent's objective (overall loss), the DP-FFL and the FFL mechanisms outperform local learning, when $\Delta\leq 0.15$. Local learning is more beneficial, compared to federated learning, when the agents have considerably non-i.i.d. data distributions (i.e., $\Delta\geq 0.15$). In particular, a higher degree of heterogeneous data implies each agent has a higher portion of  (both training and test) data samples with labels corresponding to its own index (e.g., agent $3$ may have more (both training and test) data samples with label $3$ when $\Delta$ increases). This means that individual local datasets are more ``useful''  when $\Delta$ is large, therefore, incurring a higher test accuracy for local learning.
Finally, Fig. \ref{Fig3} (a) and (b) imply whenever local learning is less beneficial than the proposed mechanisms regarding the test accuracy, the FFL and the DP-FFL schemes are more profitable for individual agents, which is consistent with our risk-bound-based voluntary participation results in Proposition \ref{P6} and Corollary \ref{C2} under Assumption \ref{Assum5}.

\textbf{Impacts of privacy risk $\alpha$ and the  number of group $L$}.
We study the impacts of the privacy risk $\alpha$ and the number of groups $L$ in Fig. \ref{Fig5}. Note that the manipulated FedAvg and the proposed FFL algorithm are not directly comparable here, as they cannot guarantee differential privacy.
We set $\Delta=0.05$ and $K=10$. Fig. \ref{Fig5} (a) shows that both the mean and the standard deviation of the payment accuracy loss decrease in $\alpha$, as a larger privacy risk $\alpha$ leads to less noise in the DP-FFL algorithm. An interesting observation is that, to attain a reasonably small payment accuracy loss, one should choose a small $T_2$ for a small $\alpha$, which is consistent with Proposition \ref{P9}.
Fig. \ref{Fig5} (b) shows that increasing the number of groups $L$ reduces the payment accuracy loss. In addition,
a relatively large enough number of groups (i.e., $L\geq 4$) is enough to maintain a relative small payment accuracy loss.


\section{Conclusions} \label{Sec:Conclusion}

We have studied an economic approach to federated learning robust against strategic agents' manipulation.
We have analyzed how the key feature of federated learning, unbalanced  and non-i.i.d. data, affects agents' incentive to voluntarily participate and obediently follow federated learning algorithms.
We have designed the first faithful mechanism for federated learning, achieving (provably approximately) optimality, faithful implementation, voluntary participation, with the time complexity (in terms of the number of agents $K$) of $\mathcal{O}(\log{K})$.
We have further presented the differentially private faithful federated learning mechanism, which is the first differentially private faithful mechanism.
It provides scalability, maintains the economic properties, and enables one to make three-way performance tradeoffs among privacy, convergence, and payment accuracy loss.

There are a few future directions. First, we assume that the (energy) cost of computation and communication is negligible. It is important to consider and analyze the impacts of such cost, and design economic mechanisms that is not only faithful but also elicits the right amount of efforts. 
Second, it is also interesting to design faithful algorithms and corresponding economic mechanisms for other more sophisticated federated learning architectures (e.g.,
multi-task federated learning \cite{FL7}).



\begin{IEEEbiography}[{\includegraphics[width=1in,height=1.25in,clip,keepaspectratio]{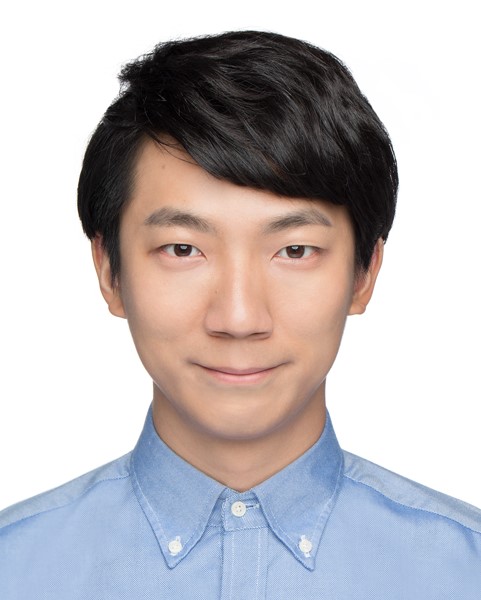}}]{Meng Zhang} (S'15 -- M'19) 
is an Assistant Professor with the Zhejiang University/University of Illinois at Urbana-Champaign Institute (ZJU-UIUC Institute), Zhejiang University.
He has been a Postdoctoral Fellow with the Department of Electrical and Computer Engineering at Northwestern University from 2020 to 2021. He received his Ph.D. degree in Information Engineering from the Chinese University of Hong Kong in 2019. He was a visiting student research collaborator with the Department of Electrical Engineering at Princeton University  from 2018 to 2019. His primary research interests include network economics and wireless networks, with a current emphasis on mechanism design and optimization for age of information and federated learning.
\end{IEEEbiography}



\begin{IEEEbiography}[{\includegraphics[width=1in,height=1.25in,clip,keepaspectratio]{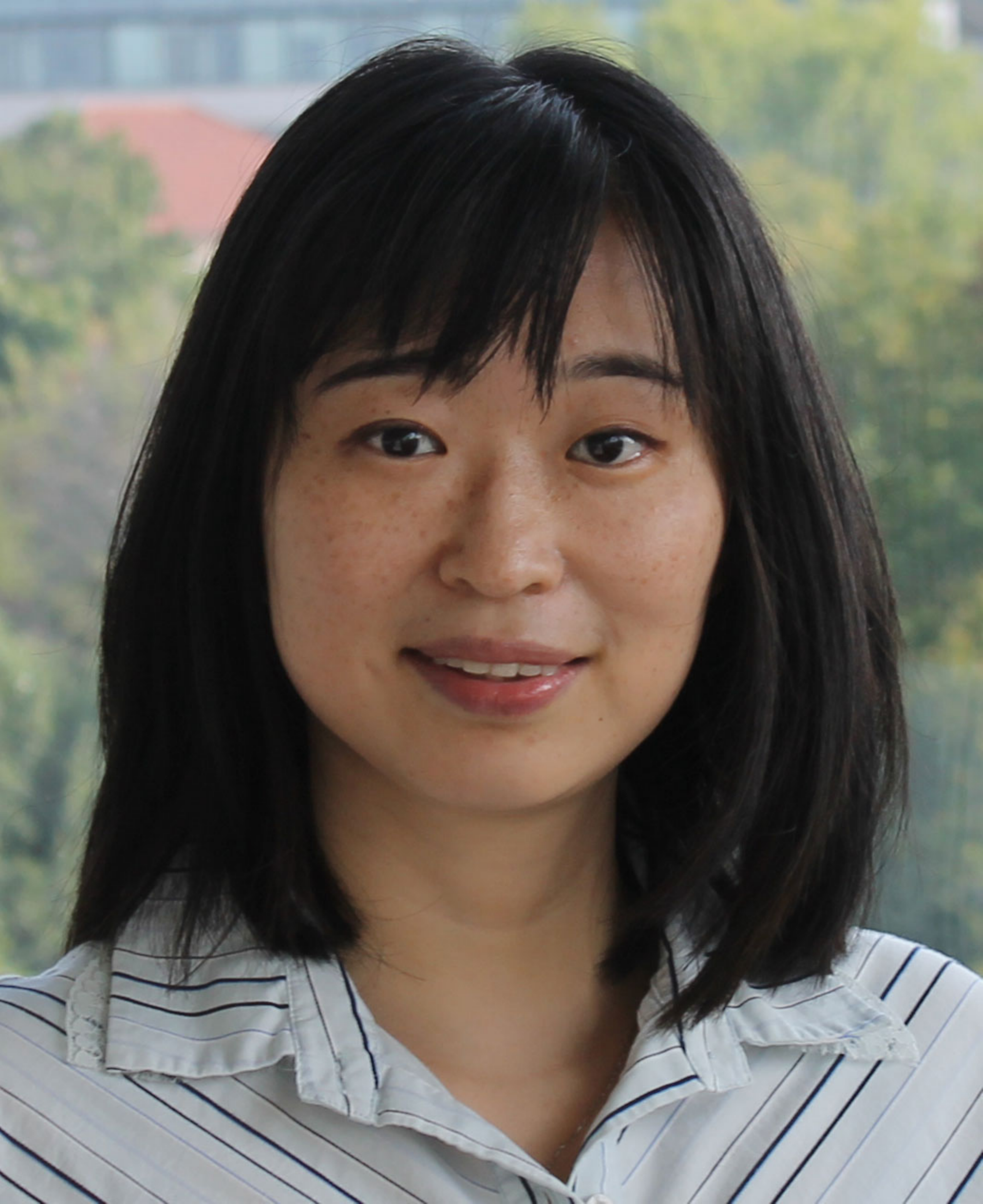}}]{Ermin Wei} is currently an Assistant Professor at the Electrical and Computer Engineering Department and Industrial Engineering and Management Sciences Department of Northwestern University. She completed her PhD studies in Electrical Engineering and Computer Science at MIT in 2014, advised by Professor Asu Ozdaglar, where she also obtained her M.S.. She received her undergraduate triple degree in Computer Engineering, Finance and Mathematics with a minor in German, from University of Maryland, College Park. Wei has received many awards, including the Graduate Women of Excellence Award, second place prize in Ernst A. Guillemen Thesis Award and Alpha Lambda Delta National Academic Honor Society Betty Jo Budson Fellowship. Her team also won the 2nd place in the Grid Optimization (GO) competition 2019, an electricity grid optimization competition organized by Department of Energy. Wei's research interests include distributed optimization methods, convex optimization and analysis, smart grid, communication systems and energy networks and market economic analysis.
\end{IEEEbiography}

\begin{IEEEbiography}[{\includegraphics[width=1in,height=1.25in,clip,keepaspectratio]{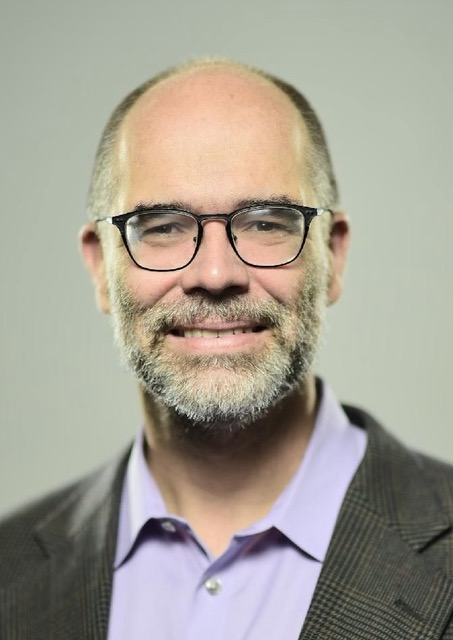}}]{Randall Berry} (F’14)
 is the John A. Dever Professor and Chair of Electrical and Computer Engineering at Northwestern University. He is also a Principle Engineer with Roberson and Associates and 
has been on the technical staff of MIT Lincoln Laboratory. He received the M.S. and Ph.D. degrees from the Massachusetts Institute of Technology in 1996 and 2000, respectively,
and the BS degree from the University of Missouri Rolla in 1993. Dr. Berry is the recipient of a NSF CAREER award and an IEEE Fellow. He has served as an Editor for the IEEE Transactions on Wireless Communications from 2006 to 2009, and an Associate Editor for the IEEE Transactions on Information Theory from 2009 to 2011. He is currently a Division Editor for the Journal of Communications and Networks and an Area editor for the
IEEE Open Journal of the Communications Society. He has also been a guest editor for special issues of the IEEE Journal on Selected Topics in Signal Processing, the IEEE Transactions on Information Theory and the IEEE Journal on Selected Areas in Communications. He has served on the program and organizing committees of numerous conferences including serving as a chair of the 2012 IEEE Communication Theory Workshop, a TPC chair of 2010 IEEE ICC Wireless Networking Symposium, and a TPC chair of the 2018 ACM Mobihoc conference.

\end{IEEEbiography}

\title{Technical Report of ``Faithful Edge Federated Learning: Scalability and Privacy''}

\maketitle

\section{Technical Report of ``Faithful Edge Federated Learning: Scalability and Privacy''}

\subsection{Proof of Proposition \ref{P1} and  Corollary \ref{C1}}\label{AL1}

In this appendix, we prove Proposition \ref{P1}, which also proves Corollary \ref{C1} as  a special case ($p_k=1$ and $p_j=0$ for all $j\neq k$).


We start with the following the Hoeffding's Inequality. Consider independent $X_i\in[a_i,b_i]$ and their sum, $S_n=\sum_{i=1}^n X_i$. The Hoeffding's Inequality states that
\begin{align}
    {\rm Pr}(S_n-\mathbb{E}[S_n]\geq \epsilon)\leq \exp\left(\frac{-2\epsilon^2}{\sum_{i=1}^n(b_i-a_i)^2}\right).\label{ProofL2Eq1}
\end{align}
The optimality condition of \eqref{Eq-P1} is
\begin{align}
    \sum_{k\in\mathcal{K}}p_k \nabla_{\bs{w}} F_k(\bs{w}^o)=\bs{0}.\label{ProofL2Eq2}
\end{align}
Since $\mathbb{E}[F_k(\bs{w})]={E}_k(\bs{w})$ for all $\bs{w}$, it follows that 
\begin{align}
    \mathbb{E}[\nabla_{\bs{w}}F_k(\bs{w}^o)]=\nabla_{\bs{w}}{E}_k(\bs{w}^o),~\forall k\in\mathcal{K}.
\end{align}

Let ${w}_l$ be the $l$-th entry of $\bs{w}$. Recall that $L_f=\max_{(\bs{x},y)\in\mathcal{D}} \norm{\nabla_{\bs{w}}\ell(\bs{w}^{o},\bs{x},y)}_2$.
For a given $l$, we have
\begin{align}
    &~~~{\rm Pr}\left(\left|\frac{\partial F(\bs{w}^o)}{\partial w_l}- \frac{\partial E(\bs{w}^o)}{\partial w_l}\right|> \epsilon\right)\nonumber\\
    &= {\rm Pr}\left(\left| \frac{\partial E(\bs{w}^o)}{\partial w_l}\right|> \epsilon\right)\nonumber\\
    &\leq 2\exp\left(\frac{-2\epsilon^2}{L_f^2\sum_{k\in\mathcal{K}}p_k^2/n_k}\right).
\end{align}
 it follows that
\begin{align}
    &{\rm Pr}\left(\exists l\in\{1,...,d\}~{\rm s.t.}~\left| \frac{\partial E(\bs{w}^o)}{\partial w_l}\right|> \epsilon\right)\nonumber\\
    =~ & {\rm Pr}\left(\left(\left| \frac{\partial E(\bs{w}^o)}{\partial w_1}\right|> \epsilon\right) \cup ...\cup \left(\left| \frac{\partial E(\bs{w}^o)}{\partial w_d}\right|> \epsilon\right)\right)\nonumber\\
    \leq~&\sum_{l=1}^{d} {\rm Pr}\left(\left| \frac{\partial E(\bs{w}^o)}{\partial w_l}\right|> \epsilon\right)\nonumber\\
    \leq ~&2d \exp\left(\frac{-2\epsilon^2}{L_{\ell}^2\sum_{k\in\mathcal{K}}p_k^2/n_k}\right).
\end{align}

By setting $\delta=2d \exp\left(\frac{-2\epsilon^2}{L_{\ell}^2\sum_{k\in\mathcal{K}}p_k^2/n_k}\right)$, we have that, with a probability of $1-\delta$, the following inequality holds:
\begin{align}
      -\sqrt{\sum_{k\in\mathcal{K}}\frac{p_k^2}{n_k}\frac{L_f^2\log(2d/\delta)}{2}} \bs{1}\preceq &\nabla_{\bs{w}}{E}_k(\bs{w}^o)\nonumber\\
      \preceq  &\sqrt{\sum_{k\in\mathcal{K}}\frac{p_k^2}{n_k}\frac{L_f^2\log(2d/\delta)}{2}} \bs{1}\nonumber\\
    \iff \quad  \norm{\nabla_{\bs{w}}{E}(\bs{w}^o)}_2^2\leq & \sum_{k\in\mathcal{K}}\frac{p_k^2}{n_k}\frac{L_f^2d\log(2d/\delta)}{2},
\end{align}
where $\bs{1}$ denotes an all-ones vector.
By strict convexity of $E(\bs{w})$, we have that,  with a probability of $1-\delta$,  the following inequality holds: 
\begin{align}
    E(\bs{w}^o)-\min_{\bs{w}} E(\bs{w})\leq &
    \frac{1}{2\mu}\norm{\nabla_{w} E(\bs{w}^o)}_2^2\nonumber\\
    \leq & \sum_{k\in\mathcal{K}}\frac{p_k^2}{n_k}\frac{L_f^2d\log(2d/\delta)}{4\mu}.
\end{align}
We complete the proof of Proposition \ref{P1}. By replacing $\bs{w}^o$ by $\bs{w}^L_k$ and ${E}(\cdot)$ by ${E}_k(\cdot)$
and setting $p_k=1$ and $p_j=0$ for all $j \neq k$, we also complete the proof of Corollary \ref{C1}.

\subsection{Proof of Proposition \ref{P2}}\label{AP2}
Define $\lambda\in[0,1]$ and $\tilde{E}(\bs{w},\lambda)$ such that
\begin{align}
    \tilde{E}_k(\bs{w},\lambda)=(1-\lambda)\sum_{j\in\mathcal{K}}p_j E_j(\bs{w})+\lambda E_k(\bs{w}).
\end{align}
We have $\tilde{E}_k(\bs{w},1)=E_k(\bs{w})$ and $\tilde{E}_k(\bs{w},0)=E(\bs{w})$ for all $\bs{w}$. Define
\begin{align}
\bs{w}_k^*(\lambda)&\triangleq\arg\min_{\bs{w}}\tilde{E}_k(\bs{w},\lambda),~\forall k\in\mathcal{K},\\
   V_k(\lambda)&\triangleq\min_{\bs{w}}\tilde{E}_k(\bs{w},\lambda),~\forall k\in\mathcal{K}.
\end{align}
By the envelope theorem \cite{Envelope},
\begin{align}
    \frac{\partial V_k(\lambda)}{\partial \lambda}= &\frac{\partial \tilde{E}_k(\bs{w}_k^*(\lambda),\lambda)}{\partial \lambda}\nonumber\\
    = &\int \ell (\bs{w}^*_k(\lambda))dP_k-\sum_{j\in\mathcal{K}}\int \ell (\bs{w}^*_k(\lambda))p_j dP_j.
\end{align}
Therefore, we have
\begin{align}
     &\left|\min_{\bs{w}}E(\bs{w})- \min_{\bs{w}}E_k(\bs{w})\right|\nonumber\\
     =&\left|\int_{0}^1 \frac{\partial \tilde{E}_k(\bs{w}_k^*(\lambda),\lambda)}{\partial \lambda}d \lambda\right|\nonumber\\
     =&\left|\int_{0}^1 \int \ell (\bs{w}^*_k(\lambda))\left(dP_k-\sum_{j\in\mathcal{K}}p_j dP_j\right) d \lambda\right|\nonumber\\
     \leq &\int_{0}^1 \left|\int \ell (\bs{w}^*_k(\lambda))\left(dP_k-\sum_{j\in\mathcal{K}}p_j dP_j\right)\right| d \lambda\nonumber\\
     \leq & \norm{P_k(\cdot)-\sum_{j\in\mathcal{K}}p_j P_j(\cdot)}.\label{Proof-P2-Eq1}
\end{align}
In addition, we have
\begin{align}
     |E(\bs{w}^o)- E_k(\bs{w}^o)|=&\left|\int \ell(\bs{w}^o) dP_k-
     \sum_{j\in\mathcal{K}}p_j\int \ell(\bs{w}^o) dP_j\right|\nonumber\\
     \leq & \norm{P_k(\cdot)-\sum_{j\in\mathcal{K}}p_j P_j(\cdot)}.\label{Proof-P2-Eq2}
\end{align}
Combining \eqref{Proof-P2-Eq1}, \eqref{Proof-P2-Eq2}, and Proposition \ref{P1}, we complete the proof of Proposition \ref{P2}.



\subsection{Proof of Proposition \ref{P3}}\label{AP3}
The problem in \eqref{MFL} is equivalent to 
\begin{align}
    \min_{\bs{w}} \sum_{k\in\mathcal{K}}\tilde{p}_kF_k(\bs{w}),\label{Proof-P3}
\end{align}
where $\tilde{p}_k=\frac{\beta p_k}{1+(\beta-1)p_k}$ and $\tilde{p}_j=\frac{p_j}{1+(\beta-1)p_k}$. Since \eqref{Proof-P3} is similar to the FL problem in \eqref{FL}, replacing  $p_k$ by $\tilde{p}_k$ for all $k\in\mathcal{K}$ in Proposition \ref{P2} proves Proposition \ref{P3}.

\subsection{Proof of Proposition \ref{P3.5}}\label{ProofP4}
From Lemma \ref{L1}, we have that
\begin{align}
   \frac{1}{2L_g}\norm{\nabla F(\bs{w}^*)}_2^2\leq  & ~F(\bs{w}^*)- F(\bs{w}^o)\nonumber\\
   \leq &\left(1-\frac{\mu}{L_g}\right)^{T_1} ( F(\bs{w}[0])-F(\bs{w}^o)). \label{PP4-Eq1}
\end{align}
As in the proof of Proposition \ref{P1}, we have
\begin{align}
    {\rm Pr}\left(\left|\frac{\partial F(\bs{w}^*)}{\partial w_l}- \frac{\partial E(\bs{w}^*)}{\partial w_l}\right|> \epsilon\right)
    \leq 2\exp\left(\frac{-2\epsilon^2}{L_{\ell}^2\sum_{k\in\mathcal{K}}p_k^2/n_k}\right),
\end{align}
based on which we can prove Proposition \ref{P3.5}.

By setting $\delta=2d \exp\left(\frac{-2\epsilon^2}{\sum_{k\in\mathcal{K}}p_k^2/n_k}\right)$, we have that, with a probability of $1-\delta$, the following inequality holds:
\begin{align}
      \sqrt{\sum_{k\in\mathcal{K}}\frac{p_k^2}{n_k}\frac{L_g^2\log(2d/\delta)}{2}} \bs{1}&\preceq\nabla_{\bs{w}}{E}(\bs{w}^*)-\nabla_{\bs{w}}{F}(\bs{w}^*)\nonumber\\
      &\preceq  \sqrt{\sum_{k\in\mathcal{K}}\frac{p_k^2}{n_k}\frac{L_{\ell}^2\log(2d/\delta)}{2}} \bs{1}.
      \end{align}
      We have
\begin{align}
&~~
\norm{\nabla_{\bs{w}}{E}(\bs{w}^*)}_2^2\nonumber\\
     &\overset{(a)}{\leq} \sum_{k\in\mathcal{K}}\frac{p_k^2}{n_k}L_\ell^2d\log(2d/\delta)+2\norm{\nabla_{\bs{w}}{F}(\bs{w}^*)}_2^2\nonumber\\
    & \overset{(b)}{\leq} \sum_{k\in\mathcal{K}}\frac{p_k^2}{n_k} L_\ell^2d\log(2d/\delta),\nonumber\\
    &~~~+4L_g\left(1-\frac{\mu}{L_g}\right)^{T_1} ( F(\bs{w}[0])-F(\bs{w}^o)),
\end{align}
where $(a)$ results from $\norm{\bs{x}+\bs{y}}^2\leq \norm{\bs{x}+\bs{y}}^2+\norm{\bs{x}-\bs{y}}^2=2\norm{\bs{x}}^2+2\norm{\bs{y}}^2$ for any two vectors, and $(b)$ is from  \eqref{PP4-Eq1}.
By strict convexity of $E(\bs{w})$, we have that,  with a probability of $1-\delta$,  the following inequality holds:
\begin{align}
    E(\bs{w}^*)-\min_{\bs{w}} E(\bs{w})&\leq  
    \frac{1}{2\mu}\norm{\nabla_{w} E(\bs{w}^*)}_2^2\nonumber\\
    &\leq \sum_{k\in\mathcal{K}}\frac{p_k^2}{n_k}\frac{L_{\ell}^2d\log(2d/\delta)}{2\mu}\nonumber\\
    &~~+\frac{2L_g}{\mu}\left(1-\frac{\mu}{L_g}\right)^{T_1} ( F(\bs{w}[0])-F(\bs{w}^o)).
\end{align}


\subsection{Proof of Proposition \ref{P4}}\label{AP4}

In this proof, we introduce the following  definitions:
\begin{align}
    \Delta \bs{w}_{-k}[t]\triangleq &~ \bs{w}_{-k}[t+1]-\bs{w}_{-k}[t]\nonumber\\= &~
    \eta_2 \sum_{j\neq k}\nabla F_j(\bs{w}_{-k}[t]),~\forall k\in\mathcal{K}, t\in\{0,1,...,T_2\},
\end{align}
\begin{align}
    \bs{G}_{-k}(\bs{w})\triangleq\sum_{j\neq k} \frac{p_j}{p_k}\nabla {F}_{j}(\bs{w})\leq \frac{1-p_k}{p_k}L_f, ~\forall k\in\mathcal{K},
\end{align}
and
\begin{align}
    g_{k,t}(x)=&x \bs{G}_{-k}(\bs{w}[t])^T\Delta\bs{w}[t]\nonumber\\
    &-\int_{0}^x \bs{G}_{-k}(\bs{w}[t]+z\cdot \Delta \bs{w}[t])^T \Delta \bs{w}[t] dz,
\end{align}
for all $k\in\mathcal{K}, t\in\{0,1,...,T_2\}$.
It follows that
\begin{align}
 \frac{\partial g_{k,t}}{\partial x}=&~\bs{G}_{-k}(\bs{w}[t])^T\Delta  \bs{w}[t]- \bs{G}_{-k}(\bs{w}[t]+x\cdot \Delta \bs{w}[t])^T \Delta \bs{w}[t]\nonumber\\
\leq &~(\bs{G}_{-k}(\bs{w}[t]) - \bs{G}_{-k}(\bs{w}[t]+x\cdot \Delta \bs{w}[t]))^T \Delta\bs{w}[t] \nonumber\\
\leq &~\frac{1-p_k}{p_k}L_gx\norm{\Delta \bs{w}[t]}_2^2.
\end{align}

We have
\begin{align}
    \sum_{t=1}^{T_2} \norm{\Delta \bs{w}[t]}_2^2&=\sum_{t=1}^{T_2} \norm{\eta_2p_k\bs{G}_{-k}[t] }_2^2\nonumber\\
     &\leq L_f^2 T_2 \eta_2^2.
\end{align}





Collectively,  we have
\begin{align}
    \left|\mathcal{P}_k^* - \sum_{j\neq k}\frac{p_j}{p_k}\left[F_j(\bs{w}^*)-F_j(\bs{w}_{-k}^*)\right]\right|& 
    \leq 
    \sum_{t=1}^T g_{k,t}(1)\nonumber\\
    &\leq \frac{1-p_k}{p_k}L_g  \sum_{t=1}^{T_2} \norm{\Delta \bs{w}[t]}_2^2\nonumber\\
    & \leq \frac{1-p_k}{p_k}L_g L_f^2T_2 \eta_2^2,
\end{align}
which completes the proof.

\subsection{Proof of Lemma \ref{L5}}\label{Proof-L2}

Define $\bs\lambda =\{\lambda_k\}_{k\in\mathcal{K}}$ and $\lambda_k\in[0,1]$.
Define a vector function 
\begin{align}
    \mathcal{F}(\bs{w},\bs\lambda)=\sum_{k\in\mathcal{K}}\lambda_k p_k\nabla_{\bs{w}} F_k(\bs{w}).
\end{align}
Define $\bs{w}^*(\bs\lambda)$ to be the solution to the following implicit function:
\begin{align}
    \mathcal{F}(\bs{w},\bs\lambda)=\sum_{k\in\mathcal{K}}\lambda_k p_k\nabla_{\bs{w}} F_k(\bs{w})=\bs{0},~\forall \lambda_k\in[0,1].\label{z12}
\end{align}
Note that $\bs{w}_{-k}^*$ in \eqref{-k} is the same as $\bs{w}^*(\hat{\bs\lambda})$
given $\hat{\bs\lambda}$ such that $\hat{\lambda}_k=0$ and $\hat{\lambda}_j=1$ for all $j\neq k$.
By the implicit function theorem, we have 
\begin{align}
   \left[\frac{\partial w_k}{\partial \lambda_j}\right]_{k,j}=-(\nabla_{\bs{w}} \mathcal{F}(\bs{w},\bs\lambda))^{-1}\cdot \nabla_{\bs\lambda}\mathcal{F}(\bs{w},\bs{\lambda}),\label{Proof-L2-Eq1}
\end{align}
where 
\begin{align}
    \nabla_{\bs{w}} \mathcal{F}(\bs{w},\bs\lambda)=\sum_{k\in\mathcal{K}}\lambda_kp_k\nabla^2_{\bs{w}}F_k(\bs{w}),
\end{align}
and
\begin{align}
    \nabla_{\bs{\lambda}} \mathcal{F}(\bs{w},\bs\lambda)=\begin{pmatrix}p_1\nabla F_1(\bs{w})& p_2\nabla F_2(\bs{w})& \cdot\cdot\cdot & p_K\nabla F_K(\bs{w})\end{pmatrix}.
\end{align}
Note that,  $F_k(\bs{w})$ is positive definite due to the strong convexity of $F_k(\bs{w})$. Therefore, $(\nabla_{\bs{w}} \mathcal{F}(\bs{w},\bs\lambda))^{-1}$ is well defined.

Equation \eqref{Proof-L2-Eq1} can be re-expressed as
\begin{align}
    \frac{\partial \bs{w}^*(\bs\lambda)}{\partial \lambda_k}=-(\nabla_{\bs{w}} \mathcal{F}(\bs{w},\bs\lambda))^{-1} p_k \nabla F_k(\bs{w}),~\forall k\in\mathcal{K}.
\end{align}
The distance between $\bs{w}^o$ and $\bs{w}^*_{-k}$ is then bounded by, where $\lambda_j=1$  for all $j\neq k$ and $p_k=\frac{1}{K}$ for all $k$,
\begin{align}
    &~~\norm{\bs{w}^o-\bs{w}^*_{-k}}_2\nonumber\\
    &=\norm{\int_{0}^{1} \frac{\partial \bs{w}^*(\bs\lambda)}{\partial \lambda_k} d {\lambda_k}}_2\nonumber\\
     &\leq \int_{0}^{1} \norm{\frac{\partial \bs{w}^*(\bs\lambda)}{\partial \lambda_k}}_2 d {\lambda_k}\nonumber\\
     &\leq \int_{0}^{1} \norm{\left(\sum_{k\in\mathcal{K}}\nabla_{\bs{w}}^2 F_k(\bs{w},\bs\lambda)\right)^{-1} \nabla F_k(\bs{w})}_2 d {\lambda_k}\nonumber\\
      &\leq \int_{0}^{1} \norm{\left(\sum_{k\in\mathcal{K}}\nabla_{\bs{w}}^2 F_k(\bs{w},\bs\lambda)\right)^{-1}}_2\norm{ \nabla F_k(\bs{w})}_2 d {\lambda_k}\nonumber\\
       &\overset{(a)}{\leq} \int_{0}^{1} \frac{L_f }{\mu K}d {\lambda_k}= \frac{L_f}{\mu K},
\end{align}
where $(a)$ is from $\mu$-strong convexity (Assumptions \ref{Assum1}) and Definition \ref{DefAsum2}.


\subsection{Proof of Theorem \ref{T1}}\label{PT1}




\begin{lemma}\label{L4}
In Algorithm \ref{Algo1}, the following inequality holds:
\begin{align}
    \norm{\bs{w}[t]-\bs{w}^o}_2^2\leq (1-\mu \eta)^{t} \norm{\bs{w}[0]-\bs{w}^o}_2^2, \forall t\in\{0,1,...,T_1\}.
\end{align}
 
\end{lemma}

From Lemma \ref{L4}, we have that the distance between the exact solution and the output solution is $\norm{\bs{w}^o-\bs{w}^*}_2\leq G\sqrt{\left(1-\frac{\mu}{L_g}\right)^{T_1}}$, where $G\triangleq\norm{\bs{w}[0]-\bs{w}^o}_2$.
We set $T_1$ such that
\begin{align}
    G\sqrt{\left(1-\frac{\mu}{L_g}\right)^{T_1}}\leq \frac{\Delta}{K},
\end{align}
which is equivalent to
\begin{align}
 T_1\geq \frac{2\ln\left(\frac{KG}{\Delta}\right)}{\ln \left((1-\frac{\mu}{L_g})^{-1}\right)}.
\end{align}

From Lemma \ref{L5}, we have 
\begin{align}
\norm{\bs{w}_{-k}^o-\bs{w}^*}_2^2\leq & (\norm{\bs{w}_{-k}^o-\bs{w}^o}_2+\underbrace{\norm{\bs{w}^o-\bs{w}^*}_2}_{\text{Suboptimality in Phase I}})^2\nonumber\\
\leq & \left(\frac{L_f}{\mu K} +\frac{\Delta}{K}\right)^2,\forall k\in\mathcal{K}.
\end{align}
Based on a similar argument as in Lemma \ref{L4}, we have
\begin{align}
               \norm{\bs{w}_{-k}[t]-\bs{w}_{-k}^o}_2^2&\leq \left(1-\frac{\mu}{L_g}\right)^{t} \norm{\bs{w}^*-\bs{w}_{-k}^o}_2^2\nonumber\\
               &\leq \left(1-\frac{\mu}{L_g}\right)^{t} \left(\frac{L_f}{\mu K}+\frac{\Delta}{K}\right)^2,~\forall t\in\mathbb{N}.\label{Proof-L3-Eq1}
\end{align} 
By $L_g$-smoothness, we have $$\sum_{j\neq k}\left(F_j(\bs{w}_{-k}[t])-F_j(\bs{w}_{-k}^*)\right)\leq \frac{L_g(K-1)}{2} \norm{\bs{w}_{-k}[t]-\bs{w}_{-k}^*}_2^2.$$
Combining \eqref{Proof-L3-Eq1} and Proposition \ref{P4}, we have
\begin{align}
              |\mathcal{P}_k[T_2]-\mathcal{P}_k^{\rm VCG}|
             \leq &\frac{L_g(K-1)}{2} \left(1-\frac{\mu}{L_g}\right)^{T_2} \left(\frac{L_f}{\mu K}+\frac{\Delta}{K}\right)^2\nonumber\\ & + (K-1)L_g L_f^2\left(\sum_{t=1}^{T_2} \eta^2[t] +\eta^2[1]\right).
\end{align} 
By selecting $\eta_2=\frac{1}{KL_g}$ and 
\begin{align}
    T_2\in\left[\ln\left( \frac{(L_f+\Delta\mu)^2 L_g}{\mu^2K \epsilon}
    \right)/\ln\left(\frac{L_g}{L_g-\mu}\right),\frac{\epsilon L_g K}{2 L_f^2}\right],
\end{align}
we have  $|\mathcal{P}_k[T_2]-\mathcal{P}_k^{\rm VCG}|\leq \epsilon,~\forall k\in\mathcal{K}$, and \begin{align}
    KT_2\leq \left( 1+\frac{(L_f+\Delta\mu)^2 L_g}{2\mu^2 e\epsilon}
    \right)/\ln\left(\frac{L_g}{L_g-\mu}\right)=\mathcal{O}(1).
\end{align}
We completed the proof.

\subsection{Proof of Proposition \ref{P5}}\label{AP5}
Let $A_k$ be an arbitrary action chosen by every agent $k\in\mathcal{K}$. Given   an arbitrary action $A_k$ played by agent $k$, let $\bs{w}'$ be the solution in Phase I and $\{\bs{w}'_{-k}\}_{k\in\mathcal{K}}$ be the solution in Phase II of Algorithm \ref{Algo1}, respectively,
and $\mathcal{P}_k'$ be the agent $k$'s payment for all $k\in\mathcal{K}$.



Recall that one of the termination criteria for Phase II in
Algorithm \ref{Algo1} is:
\begin{align}
  \frac{1}{2\mu }\norm{\sum_{j\neq k} p_j \nabla  F_j(\bs{w}_{-k}[t])}_2^2\leq p_k \epsilon.\label{Ke}
\end{align}
Such a termination criteria ensures that, no matter what action $A_k$ agent $k$ plays, as long as all other agents are faithful (i.e., following the intended algorithm), we must have 
\begin{align}
    \sum_{j\neq k} \frac{p_j}{p_k}\left(F_j(\bs{w}_{-k}')-F_j(\bs{w}_{-k}^*)\right)\leq \epsilon, \label{Eq-105}
\end{align}
from the $\mu$-strong convexity of $F_j$ for all $j\in\mathcal{K}$.


On the other hand, define $\mathcal{P}_k^{\rm VCG,II}\triangleq \sum_{j\neq k}\left(F_j(\bs{w}')-F_j(\bs{w}'_{-k}[T_2])\right)$ as the VCG payment the algorithm approximates when agent $k$ plays $A_k$. We have
\begin{claim}
The following inequality is true:
  \begin{align}
\mathcal{P}_k^{\rm VCG,II}
&=\sum_{j\neq k}\left(F_j(\bs{w}'_{-k}[0])-F_j(\bs{w}'_{-k}[T_2])\right)\nonumber\\
&=\sum_{t=1}^{T_2}\sum_{j\neq k}\left(F_j(\bs{w}'_{-k}[t-1])-F_j(\bs{w}'_{-k}[t])\right)\nonumber\\
&\overset{(a)}{\leq} \sum_{t=1}^{T_2} \sum_{j\neq k}\nabla F_j(\bs{w}'_{-k}[t-1])^T(\bs{w}'_{-k}[t-1]-\bs{w}'_{-k}[t])\nonumber\\
&= \mathcal{P}_k',\label{Za2}
\end{align} 
where $(a)$ is due to the convexity of $F_j(\bs{w})$ for all $\bs{w}$ and $j\in\mathcal{K}$.
\end{claim}


We are ready to focus on agent's overall cost when it plays its faithful action $s_k^m$. From $|\mathcal{P}_k[T_2]-\mathcal{P}_k^{\rm VCG}|\leq \epsilon$ as indicated in Theorem \ref{T1}, we have
\begin{align}
   \mathbb{E}[J_k({s}_k^m,\bs{s}_{-k}^m)]&\leq \mathbb{E}\left[\sum_{j\in\mathcal{K}}F_j(\bs{w}^*)-\sum_{j\neq k}F_j(\bs{w}_{-k}^*)+\epsilon\right].
   \end{align}
 From   Proposition \ref{P3.5}, it follows that, with a probability of $1-\tilde{\delta}$, the following inequality holds: for all possible $A_k$,
   \begin{align}
  &~~~\mathbb{E}[J_k({s}_k^m,\bs{s}_{-k}^m)] \nonumber\\
  &\leq  K \min_{\bs{w}} E(\bs{w})-\mathbb{E}\left[\sum_{j\neq k}F_j(\bs{w}_{-k}^*)\right]+\epsilon+K\Phi(\delta)\nonumber\\
   &\overset{(b)}{\leq} K E(\bs{w}')-\mathbb{E}\left[\sum_{j\neq k}F_j(\bs{w}_{-k}')\right]+2\epsilon+ K\Phi(\delta)\nonumber\\
   &\overset{(c)}{\leq} E_k(\bs{w}')+\mathbb{E}[\mathcal{P}_k']+2\epsilon+ K\Phi(\delta)\nonumber\\
   &= \mathbb{E}[J_k(A_k,\bs{s}_{-k}^m)]+2\epsilon+ K\Phi(\delta),
\end{align}
where (b) is from \eqref{Eq-105}, and (c) is from \eqref{Za2}.


\subsection{Proof of Proposition \ref{P6}}\label{AP6}
From Assumption \ref{Assum5}, we have that ${\rm RB}_{\{\mathcal{K}\}}\leq {\rm RB}_{\mathcal{C}}$. If we select $\mathcal{C}=\{\{k\},\mathcal{K}\backslash\{k\}\}$, the inequality ${\rm RB}_{\{\mathcal{K}\}}\leq {\rm RB}_{\mathcal{C}}$ becomes:
\begin{align}
&p_k\frac{L_{\ell}^2d\log(2d/\delta)}{4\mu n_k}+\sum_{j\neq k}\frac{p_{j}^2}{(1-p_k)^2n_k}\frac{L_{\ell}^2d\log(2Kd/\delta)}{4\mu}
\nonumber\\
&+\sum_{j\neq k}2p_j\norm{P_j(\cdot)-\bar{P}_{\mathcal{K}\backslash\{k\}}(\cdot)}\nonumber\\
\geq &\sum_{j\in\mathcal{K}}\frac{p_j^2}{n_j}\frac{L_{\ell}^2d\log(2dK/\delta)}{4\mu}
+2\sum_{k\in\mathcal{K}}p_k\norm{P_k(\cdot)-\sum_{j\in\mathcal{K}}p_j P_j(\cdot)}. \label{P6-Eq-1}
\end{align}
In addition, from $|\mathcal{P}_k[T_2]-\mathcal{P}_k^{\rm VCG}|\leq \epsilon$ as indicated in Theorem \ref{T1},
it follows that
\begin{align}
     \mathbb{E}[J_k({s}_k^m,\bs{s}_{-k}^m)]\leq \min_{\bs{w}}\sum_{j\in\mathcal{K}}\frac{p_j}{p_k}E_j(\bs{w})-\min_{\bs{w}}\sum_{j\neq k}\frac{p_j}{p_k}E_j(\bs{w})+\epsilon,  \label{P6-Eq-2}
\end{align}
 Therefore, combining \eqref{P6-Eq-1} and \eqref{P6-Eq-2}, we prove Proposition \ref{P6}.

\subsection{Proof of Proposition \ref{P7}}\label{AP7}
From \eqref{payment-1} and \eqref{payment-2}, we have
\begin{align}
    \mathcal{P}_k^*=\sum_{t=1}^{T_2}\frac{\eta_2}{p_k}\norm{\sum_{j\neq k}p_j\nabla F_j(\bs{w}_{-k}[t])}_2^2\geq 0,~\forall k\in\mathcal{K}.
\end{align}
Hence, we have completed the proof of Proposition \ref{P7}.

\subsection{Proof of Theorem \ref{T3}}\label{AT3}
Based on the proof of Lemma \ref{L5}, the distance between $\bs{w}^*_{-k}$ and $\bs{w}^*_{m}$ is given by
\begin{align}
    &~~\norm{\bs{w}_{l}^o-\bs{w}^o_{-k}}_2\nonumber\\
    &=\norm{\int_{0}^{1} \frac{\partial \bs{w}^o(\bs\lambda)}{\partial \lambda_k} d {\lambda_k}}_2\nonumber\\
     &\leq \int_{0}^{1} \norm{\frac{\partial \bs{w}^*(\bs\lambda)}{\partial \lambda_k}}_2 d {\lambda_k}\nonumber\\
     &\leq \int_{0}^{1} \norm{\left(\sum_{k\in\mathcal{K}}\nabla_{\bs{w}}^2 F_k(\bs{w},\bs\lambda)\right)^{-1} \nabla F_k(\bs{w})}_2 d {\lambda_k}\nonumber\\
      &\leq \int_{0}^{1} \norm{\left(\sum_{k\in\mathcal{K}}\nabla_{\bs{w}}^2 F_k(\bs{w},\bs\lambda)\right)^{-1}}_2\sum_{j\in\mathcal{C}_l\backslash \{k\}}\norm{ \nabla F_j(\bs{w})}_2 d {\lambda_j}\nonumber\\
       &\leq \int_{0}^{1} \frac{L_f (\lceil K/M \rceil-1)}{\mu K}d {\lambda_k}\leq \frac{L_f}{\mu M}.
\end{align}
Hence, we have 
\begin{align}
    |\mathcal{P}_k^{\rm S}-\mathcal{P}_k^{\rm VCG}|&=\left|\sum_{j\neq k} \left(F_j(\bs{w}_m^*)-F_j(\bs{w}^*_{-k})\right)\right|
    \nonumber\\
    &\overset{(a)}{\leq} \frac{(K-1)L_g}{2} \norm{\bs{w}_{m}^*-\bs{w}^*_{-k}}_2^2\nonumber\\
    &\leq \frac{(K-1)L_f^2 L_g}{2\mu^2 M^2},~\forall k\in\mathcal{K},
\end{align}
where (a) is by the $L_g$-smoothness of $F_j(\bs{w})$.
Hence, if we select
\begin{align}
    M\geq \min\left\{K,\sqrt{\frac{L_g (K-1)}{2 \epsilon }}\frac{L_f}{\mu}\right\}=\mathcal{O}\left(\sqrt{\frac{K}{\epsilon}}\right).
\end{align}
then we have $\left|\mathcal{P}_k^{\rm S}-\mathcal{P}_k^{\rm VCG}\right|\leq \epsilon$.

\subsection{Proof of Proposition \ref{P8}}\label{AP8}


     The main proof of Proposition \ref{P8} is based on \cite{SDP}. We start with the following intermediate 
    notation of concentrated differential privacy \cite{concen}: 
\begin{definition}
A randomized mechanism $\mathcal{M}: \mathcal{D}^n\rightarrow \mathbb{R}^d$ is $\rho$-Zero-Concentrated Differential Privacy ($\rho$-zCDP) if for any two adjacent datasets $\mathcal{X}$ and  $\mathcal{X}'\in\mathcal{D}^n$ differing in one sample, it holds for all $a\in(1,\infty)$ that
\begin{align}
    \mathbb{E}\left[\exp\left((a-1)\log\frac{\mathbb{P}[\mathcal{M}(\mathcal{\mathcal{X}})=y]}{\mathbb{P}[\mathcal{M}(\mathcal{\mathcal{X}'})=y]}\right)\right]\leq \exp((a-1)a\rho),
\end{align}
for $y\in\mathbb{R}^d$.

\end{definition}

Bun and Steinke in \cite{concen} provided the following lemmas to achieve zCDP with the Gaussian mechanism. Lemmas \ref{LL1}-\ref{LL3} bound the amount of Gaussian noise to guarantee zCDP, give the composition of multiple zCDP mechanisms, and specify the mapping from different privacy metrics.

\begin{lemma}\label{LL1}
Given a function $q: \mathcal{D} \rightarrow \mathbb{R}^n$, the outcome  $\mathcal{M}=q(\mathcal{D})+\bs{u}$, where $\bs{u}\in \mathcal{N}(\bs{0},\sigma^2 \bs{I}_d)$, satisfies $\Delta_2(q)^2/(2\sigma^2)$-zCDP.
\end{lemma}

\begin{lemma}\label{LL2}
For any two mechanisms
$\mathcal{M}_1: \mathcal{D} \rightarrow \mathbb{R}^d$ and $\mathcal{M}_2: \mathcal{D} \times \mathbb{R}^d\rightarrow \mathbb{R}^d$. $\mathcal{M}_2(D,\mathcal{M}_1(D))$ satisfies $(\rho_1+\rho_2)$-zCDP.
\end{lemma}

\begin{lemma}\label{LL3}
 If a randomized mechanism $\mathcal{M}$: $\mathcal{D}^n\rightarrow \mathbb{R}^d$ satisfies $\rho$-zCDP, then it satisfies $(\rho+2\sqrt{\rho\log(1/\delta)},\delta)$-DP for any $\delta$.
\end{lemma}

Define $F_k(\bs{w},\mathcal{D}_k)$ as the empirical risk given the local dataset $\mathcal{D}_k$:
\begin{align}
    F_k(\bs{w},\mathcal{D}_k)=\frac{1}{K}\sum_{k=1}\frac{1}{n_k}\sum_{i=1}^{n_k} \ell(\bs{w},\bs{x}_{i},y_i),
\end{align}
Given a gradient computation in iteration $t$ in Algorithm \ref{Algo2}, we define
\begin{align}
    M_{k,t}\triangleq&\nabla F_k(\bs{w},\mathcal{D}_k)+ \mathcal{N}(\bs{0},\sigma^2 \bs{I}_d)\nonumber\\=&\frac{1}{K}\sum_{k=1}\frac{1}{n_k}\sum_{i=1}^{n_k}\nabla \ell(\bs{w}^T\bs{x}_{i},y_i) + \mathcal{N}(\bs{0},\sigma^2 \bs{I}_d).
\end{align}
Supposing $\mathcal{D}_k$ and $\mathcal{D}_k'$ only differ in one sample, we have
\begin{align}
    \norm{\nabla F(\bs{w},\mathcal{D}_k)-\nabla F(\bs{w},\mathcal{D}_k')}_2\leq \frac{2L_f}{K n_{(1)}}.\label{P10-Eq1}
\end{align}
Thus, using Lemma \ref{LL1}, we have that $M_{k,t}$ is $\rho$-zCDP where $\rho=\frac{2 L_f^2}{K^2 n_{(1)}^2 \sigma^2}$.
On the other hand, by adopting the noise $\sigma^{2}_P$ in \eqref{noise}, we have that 
\begin{align}
    M'_k&
    =\mathcal{P}_k^{S}(\mathcal{D})+\mathcal{N}({0},\sigma_P^2), \forall k\in\mathcal{K},
\end{align}
where $\mathcal{P}_k^{S}(\mathcal{D})\triangleq\sum_{j\neq k} \left({F}_j(\bs{w}^*,\mathcal{D}_j)- {F}_j(\bs{w}_{l}^*,\mathcal{D}_j)\right).$
Suppose that $\mathcal{D}$ and $\mathcal{D}'$ only differ in one sample,
\begin{align}
    \norm{\mathcal{P}_k^{S}(\mathcal{D})-\mathcal{P}_k^{S}(\mathcal{D'})}_2\leq \frac{2}{n_{(1)}}, \forall k\in\mathcal{K}.\label{P10-Eq2}
\end{align}
Using Lemma \ref{LL1}, we have that $M_{k}'$ is $\rho_P$-zCDP where $\rho_P=\frac{2}{n_{(1)}^2 \sigma_P^2}$.

From Lemma \ref{LL2}, we have that $\bs{w}^*$ is $T_1\rho$-zCDP, $\{\bs{w}_l\}^*$ are $K T_2\rho$-zCDP. From \eqref{P10-Eq1} and \eqref{P10-Eq2},
 to make sure $\{\mathcal{P}_k^*\}_{k\in\mathcal{K}}$ are $(T_1+K T_2)\rho$-zCDP, i.e., $(T_1+K T_2)\rho=K \rho_P$, it follows that
 \begin{align}
     (T_1+K T_2)\rho=\frac{2 K}{n_{(1)}^2 \sigma_P^2}&=\frac{ 2(T_1+K T_2)  L_f^2}{K^2 n_{(1)}^2 \sigma^2},\nonumber\\
   \longrightarrow
 ~~~ \sigma_P^2&= \frac{\sigma^2 K^3}{(T_1+KT_2)L_f^2},
 \end{align}
which leads to an overall  $2(T_1+K T_2)\rho$-zCDP for Algorithm \ref{Algo2}, according to Lemma \ref{LL2}.
 
Lemma \ref{LL3} shows that Algorithm \ref{Algo2} is also $(2(T_1+K T_2)\rho+2\sqrt{2(T_1+K T_2)\rho \log(1/\delta)},\delta)$-DP for any $\delta\in(0,1)$. 
It follows from Lemma \ref{LL3} that
\begin{align}
    \rho&\approx \frac{\alpha^2}{ 8(T_1+T_2 K)\log(1/\beta)},\nonumber\\
    \longrightarrow
 ~~~ \sigma^2&=\frac{16L_f^2 (T_1+KT_2)\log(1/\beta)}{K^2 n_{(1)}^2\alpha^2},\\
 \sigma_P^2&=\frac{16 K \log(1/\beta)}{ n_{(1)}^2\alpha^2}.
\end{align}

\subsection{Proof of Proposition \ref{P7.5}}\label{AP7.5}
From the $L_g$-smoothness assumption, it follows that
\begin{align}
    &~~\mathbb{E}\left[F(\bs{w}[t+1])-F(\bs{w}[t])\right]\nonumber\\
    &\leq \mathbb{E}\left[\nabla F(\bs{w}[t])^T\left(F(\bs{w}[t+1])-F(\bs{w}[t])\right)+\frac{1}{2L_g}\norm{\nabla F(\bs{w}[t]) +\bs{n}}_2^2\right]\nonumber\\
    &= -\frac{1}{2L_g}\norm{\nabla F(\bs{w}[t])}_2^2+\frac{1}{2 L_g}\mathbb{E}\norm{\bs{n}}_2^2\nonumber\\
    &\leq -\frac{\mu}{L_g}(F(\bs{w}[t])-F(\bs{w}^o))+\frac{d\sigma^2}{2L_g}.
\end{align}
The last inequality results from the $\mu$-strong convexity assumption. The above equation can be written as:
\begin{align}
    \mathbb{E}\left[F(\bs{w}[t+1])-F(\bs{w}^o)\right]&\leq  \left(1-\frac{\mu}{L_g}\right)^{T_1}(F(\bs{w}[0])-F(\bs{w}^o))\nonumber\\
    &~~+\frac{d\sigma^2}{2\mu}.
\end{align}
When $T_1=\mathcal{O}\left(\log\left(\frac{ K^2n_{(1)}^2 \alpha^2}{d L_f^2 \log(1/\beta)}\right)\right)$ and $T_2=\mathcal{O}\left(\log\left(\frac{ K^2n_{(1)}^2 \alpha^2}{d L_f^2 \log(1/\beta)}\right)\right)$, it follows that
\begin{align}
        \mathbb{E}\left[F(\bs{w}[T_1])\right]-F(\bs{w}^o)&\leq  C_1\frac{L_f^2 d \log(Kn_{(1)})\log(1/\beta)}{K^2 n_{(1)}^2\alpha^2},
\end{align}
for some constant $C_1>0$.

Finally, according to \cite{Fast}, 
\begin{align}
    E(\bs{w}[T_1])-\min_{\bs{w}}E(\bs{w})\leq &2 \left[F(\bs{w}[T_1])-F(\bs{w}^o)\right]\nonumber\\
    &+\sum_{k\in\mathcal{K}}\frac{p_k^2}{n_k}\frac{L_{\ell}^2d\log(2d/\delta)}{2\mu}.
\end{align}
Therefore, we have the following holds with a probability of at least $1-\delta$,
\begin{align}
    E(\bs{w}[T_1])-\min_{\bs{w}}E(\bs{w})\leq & ~C_1\frac{L_f^2 d \log^2(Kn_{(1)})\log(1/\delta)}{K^2 n_{(1)}^2\epsilon^2}\nonumber\\
    &+\sum_{k\in\mathcal{K}}\frac{p_k^2}{n_k}\frac{L_{\ell}^2d\log(2d/\delta)}{2\mu},
\end{align}
for some constants $C_1>0$.

\subsection{Proof of Proposition \ref{P9}}\label{AP9}
We first define
\begin{align}
    J_{-k}(\bs{w})\triangleq \sum_{j \neq k}F_j(\bs{w}), \forall \bs{w}\in\mathbb{R}^d.
\end{align}
From $L_g$-smoothness assumption, it follows that
\begin{align}
   &~~~ \mathbb{E}[J_{-k}(\bs{w}[t+1])-J_{-k}(\bs{w}[t])]\nonumber\\
   &\leq \mathbb{E}\left[\nabla J_{-k}(\bs{w}[t])^T\Delta \bs{w}[t]+\frac{\eta_2}{2}\norm{\nabla J_{-k}(\bs{w}[t])+\bs{z}}_2^2\right]\nonumber\\
    &=\mathbb{E}\left[-\eta_2 \nabla F(\bs{w}[t])^T(\nabla J_{-k}(\bs{w}[t])+\bs{z})+\frac{\eta_2}{2}\norm{\nabla J_{-k}(\bs{w}[t])+\bs{z}}_2^2\right]\nonumber\\
    &=-\frac{\eta_2}{2}\norm{\nabla J_{-k}(\bs{w}[t])}^2_2+\frac{\eta}{2}\mathbb{E}\norm{\bs{z}}^2_2\nonumber\\
    &\leq - (K-1)\mu \eta_2(J_{-k}(\bs{w}[t])-J_{-k}^*)+\frac{d \eta_2 \sigma^2}{2}.
\end{align}
Therefore, we have that
\begin{align}
    \mathbb{E}[J_{-k}(\bs{w}_l[t+1])]-J_{-k}(\bs{w}_{l}^*)\leq & \left(1-\eta_2(K-1) \mu\right)  \nonumber\\ 
    \times (J_{-k}(\bs{w}_l[t])-J_{-k}(\bs{w}_{l}^*))
    & +\frac{d\sigma^2\eta_2 }{2}\label{Proof-Prop-10-Eq}.
\end{align}
Summing over $t=0,...,T_2$ iterations, we have that
\begin{align}
    \mathbb{E}[J_{-k}(\bs{w}[t+1])]-J_{-k}(\bs{w}_{l}^*)\leq & \left(1-\eta_2 (K-1) \mu\right)^{T_2} \nonumber\\ \times (J_{-k}(\bs{w}^*)-J_{-k}(\bs{w}_{l}^*))
    & +\frac{d\sigma^2 }{2(K-1)\mu}\label{Proof-Prop-10-Eq}.
\end{align}
Note that 
\begin{align}
    J_{-k}(\bs{w}^*)-J_{-k}(\bs{w}_l^*)\overset{(a)}{\leq} \frac{(K-1) L_f^2}{2\mu}= B,
\end{align}
where $(a)$ is due to the fact that $(K-1)^2 L_f^2/2\geq 1/2 \norm{\nabla J_{-k}(\bs{w})}_2^2\geq (K-1) \mu  (J_{-k}(\bs{w})-J_{-k}(\bs{w}_{k}^*))\geq (K-1) \mu (J_{-k}(\bs{w})-J_{-k}(\bs{w}_{l}^*))$ for all $\bs{w}$ due to the strong convexity of $J_{-k}(\bs{w})$ and $\bs{w}_{k}^*=\arg \min J_{-k}(\bs{w})$.

Recall that
\begin{align}
    \sigma^2&=\frac{16L_f^2 (T_1+KT_2)\log(1/\beta)}{K^2 n_{(1)}^2\alpha^2}.
\end{align}
Collectively, the following serves as an upper bound for the right hand side of \eqref{Proof-Prop-10-Eq}:
\begin{align}
    AT_2+BC^{T_2},\label{ABC}
\end{align}
where
\begin{align}
    A&=\frac{8 \eta_2 d L_f^2 (K+1) \log(1/\beta)}{\mu K^2 (K-1) n_{(1)}^2 \alpha^2},\\
    B&= \frac{(K-1) L_f^2}{2\mu},\\
    C&=1-(K-1) \mu \eta_2.
\end{align}
In addition, the choice of $T_2$ in \eqref{Convergence11} is to minimize the value in \eqref{ABC}.




\end{document}